\documentclass[final,custom]{anthology-ch} 

\usepackage{booktabs}
\usepackage{graphicx}
\usepackage{amsmath}
\usepackage{tikz}
\usepackage{float}
\usetikzlibrary{arrows.meta}
\usepackage{hyperref}
\usepackage{enumitem}
\usepackage{placeins}

\title{Automatic Iconographic Annotation of Medieval Illuminations:\\
Comparing Generative Vision--Language Models and Supervised Discriminative Classifiers}

\author[1,2]{Vincent Jolivet}[
  orcid=0000-0003-0600-0362
]

\author[1]{Lucas Terriel}[
  orcid=0000-0002-9189-258X
]

\author[1,2]{Chahan Vidal-Gorène}[
  orcid=0000-0003-1567-6508
]

\affiliation{1}{École nationale des chartes-PSL, Paris, France}
\affiliation{2}{Centre Jean Mabillon, Paris, France}

\keywords[eng]{Computer Vision, Vision--Language Models, Hierarchical Classification, Medieval Manuscripts}

\pubyear{2026}
\pubvolume{1}
\pagestart{1}
\pageend{1}
\paperorder{1}
\conferencename{Proceedings of Conference XXX}
\conferenceeditors{Editor1 Editor2}
\doi{00000/00000}  
\customhead{Accepted for CHR 2027} 

\firstpagefootnote{\footnotesize Paper accepted for CHR 2027.\par\vspace{0.5em}\par\noindent\textcolor{newgray}{© 2026 by the authors. Licensed under \href{https://creativecommons.org/licenses/by/4.0}{Creative Commons Attribution 4.0 International (CC BY 4.0).}}}

\begin{document}

\maketitle

\begin{abstract}
The large-scale digitisation of medieval manuscripts creates a growing need for automated iconographic indexing. We evaluate two complementary paradigms for this task on 9,735 full-folio images annotated with 2,333 TIMEL concepts: generative Vision--Language Models (VLM) and supervised discriminative classifiers (SDC). The VLM experiments evaluate fine-tuning, structured JSON decoding, taxonomy-based candidates generation, and deterministic reranking. The discriminative experiments evaluate flat, hierarchy-aware, and patch-attention heads over a frozen DINOv3 encoder.
Our results highlight key functional differences: ML-Decoder (SDC) achieves the strongest overall retrieval performance (precision@1 of 0.696, MRR of 0.806), with loss calibration (Asymmetric Loss) serving as the primary performance lever. While SDCs provide superior precision and reliability within a closed vocabulary, VLMs offer strong semantic generalization and broad taxonomic coverage, particularly through structured JSON targets and verify-calibrate reranking (hierarchical F1 of 0.559). Furthermore, model behavior strongly depends on concept abstraction: SDCs excel at perceptually grounded descriptive concepts, whereas VLMs provide useful hypotheses for complex interpretative scenes. Finally, we find that enforcing explicit hierarchy within model architectures limits coverage compared to spatial feature aggregation. Overall, rather than competing approaches, SDCs and VLMs emerge as complementary components for future hybrid indexing systems combining structured retrieval with exploratory annotation assistance. As a methodological contribution, we define the soft F1 metric to provide a taxonomy-aware evaluation that credits semantically close predictions based on lowest common ancestors.
\end{abstract}

\section{Introduction} 

\paragraph{Background.} The study of medieval images is undergoing a profound transformation driven by the large-scale digitization and open dissemination of cultural heritage collections. Until recently, access to illuminated manuscripts remained a major obstacle for researchers. Today, digital libraries such as Gallica, e-codices, the Vatican Library, and the Biblissima portal provide access to hundreds of thousands of digitized manuscripts. This shift has fundamentally changed the nature of the problem: the challenge is no longer to access images, but to efficiently discover, describe, and analyze them within increasingly large-scale digital collections.

\paragraph{Motivation.} This unprecedented abundance opens new perspectives for iconographic research. Rather than relying on a handful of well-studied collections, scholars can now investigate iconographic phenomena across manuscripts preserved by numerous institutions. However, these opportunities remain constrained by the quality and consistency of the associated metadata. Most medieval images are either sparsely annotated or described using heterogeneous indexing practices, making large-scale exploration difficult. The Objet Detection for Illuminations (O.D.I.L.) project, supported by Biblissima+, addresses this challenge by developing an automated pipeline for assisting the iconographic annotation of large collections of medieval illuminations using recent advances in computer vision.

\paragraph{Challenge.} Automatic iconographic annotation is particularly challenging from a machine learning perspective. The TIMEL thesaurus adopted in this work contains more than 4,000 hierarchically organized concepts covering material objects, architectural elements, animals, historical and biblical figures, narrative scenes, and thematic categories. The task therefore corresponds to an extreme multi-label classification problem with thousands of highly imbalanced labels, many of which are represented by only a handful of training examples.

\paragraph{Scope and contribution.} Recent work on automatic annotation has followed two main directions. The first formulates the problem as a \textit{cross-modal retrieval} task by aligning images with textual descriptions of thesaurus concepts in a shared embedding space. The second treats it as a \textit{hierarchical multi-label classification} problem by explicitly exploiting the structure of the underlying taxonomy. Meanwhile, the rapid progress of large Vision-Language Models (VLM) raises the question of whether dedicated architectures remain necessary for this task. Rather than proposing yet another architecture, this paper investigates the respective strengths and limitations of these competing paradigms. We compare a recent generative Vision-Language Model with a deterministic hierarchical classifier on a corpus of medieval illuminations annotated with the TIMEL thesaurus, with the aim of better understanding their respective capabilities for automatic iconographic indexing.


\section{Related Work}

Research on the automatic assignment of iconographic concepts has primarily focused on the Iconclass thesaurus, which remains the reference standard for indexing artworks in many cultural heritage institutions. Two methodological families have progressively emerged: \textit{cross-modal retrieval} approaches and \textit{hierarchical classification} approaches.

\paragraph{Banar et al. (2021)\cite{banar2021}} were the first to formulate Iconclass annotation as a \textit{cross-modal retrieval} task rather than a conventional classification problem. Their method relies on the SAEM architecture, in which visual representations extracted by Faster R-CNN and textual representations produced by BERT are projected into a shared semantic embedding space. Artworks are represented jointly by their images and multilingual titles, whereas Iconclass concepts are represented through their textual definitions. Assigning an Iconclass code therefore amounts to retrieving the concept definition that lies closest to the embedding of a given artwork. Their experiments show that textual metadata provide stronger discriminative power than visual features, although visual information improves performance when combined with textual descriptions. The authors nevertheless identify several important limitations: the visual and textual representations remain frozen, the model does not benefit from task-specific fine-tuning, the training corpus is relatively small, and the hierarchical structure of Iconclass is ignored.

\paragraph{Banar et al. (2023)\cite{banar2023}.} These limitations are largely addressed in their next publication. While preserving the cross-modal retrieval paradigm, the authors replace the original pipeline with a fully trainable architecture based on the MCLIP vision-language model. Both visual and textual encoders are jointly fine-tuned to better align artworks with Iconclass definitions within a shared embedding space. Their experiments demonstrate substantial improvements over both the original SAEM architecture and several general-purpose vision-language models. Nevertheless, the study also highlights the persistent challenges posed by the very large number of Iconclass concepts, the scarcity of annotated examples per class, and the difficulty of transferring the model to unseen collections or zero-shot settings.

\paragraph{Springstein et al. (2024)\cite{springstein2024}} introduce a substantial methodological shift by abandoning cross-modal retrieval in favor of \textit{hierarchical multi-label classification} explicitly designed for the structure of the Iconclass taxonomy. Their framework first performs contrastive pre-training of a CLIP model using synthetic image-text pairs automatically generated from Iconclass keywords. The resulting visual encoder is then specialized through several classification architectures, including a hierarchical Transformer decoder equipped with cross-attention mechanisms that explicitly model parent-child relationships within the taxonomy. Their results demonstrate that explicitly modeling the taxonomy substantially improves classification performance, particularly for infrequent concepts, and emphasize the importance of domain-specific adaptation of vision-language models before downstream iconographic classification.

\paragraph{Thomas and Hintersteiner (2026)\cite{thomas2026}.} Recent work explores Retrieval-Augmented Generation (RAG) with large multimodal models for automatic iconographic annotation. The authors evaluate a RAG pipeline combining image retrieval with a generative Vision-Language Model to assign Iconclass concepts to medieval illuminations from the \textit{Wenceslas Bible}. Their experiments show that such pipelines can successfully transfer iconographic knowledge across different visual traditions, although performance remains sensitive to stylistic differences and still requires expert validation. The study demonstrates the potential of generative VLMs for assisting iconographic indexing without task-specific retraining.

\vspace{4mm}
\noindent\textbf{}While most approaches rely on specialized retrieval or hierarchical classification architectures trained for Iconclass, recent work has also explored Retrieval-Augmented Generation using multimodal foundation models. However, these studies remain restricted to the Iconclass taxonomy. It therefore remains unclear whether foundation models can compete with specialized architectures when applied to a different hierarchical thesaurus such as TIMEL, designed specifically for medieval illuminations.


\section{Dataset}
\label{sec:dataset}

\paragraph{The Image database.} Our experiments are conducted on the Image database, developed and curated over several decades by the AHLoMA research group (EHESS). This corpus gathers medieval images originating from numerous Western manuscripts and constitutes one of the major reference datasets for the study of medieval iconography. It offers several properties that are particularly valuable for supervised learning: annotations are produced and validated by domain experts, they are consistent across the entire collection, and they aim to provide an exhaustive description of the iconographic content present in each image.

\paragraph{TIMEL thesaurus.} The annotation process relies on the TIMEL thesaurus. Unlike Iconclass, which was designed to encompass the entire history of Western art, TIMEL was specifically developed for the iconographic indexing of medieval images. Its vocabulary follows a documentary perspective that favors annotation consistency over excessively fine semantic distinctions. As a result, TIMEL provides a more compact yet highly specialized ontology that is particularly well suited to the study of illuminated manuscripts. Its hierarchical organization comprises more than 4,000 concepts distributed across multiple levels.

\paragraph{Descriptive vs interpretative concepts.} These concepts can be divided into two broad categories. The first corresponds to \textit{descriptive} concepts, which directly characterize the visible content of an image, such as natural elements, animals, objects, or architectural structures. The second comprises \textit{interpretative} concepts, including identified historical or biblical figures, narrative scenes, subjects, and broader iconographic themes. This distinction is important because it reflects two fundamentally different levels of visual understanding. Recognizing an object is primarily a perceptual task, whereas identifying a biblical episode or a historical character requires contextual reasoning based on the composition of the scene.

\paragraph{Dataset.} For this exploratory study, we selected a subset consisting of 9,735 full-folio images associated with approximately 109,000 annotations spanning 2,333 TIMEL concepts. Each image is annotated with approximately twelve concepts on average. As commonly observed in cultural heritage datasets, the label distribution exhibits a pronounced long-tail pattern: nearly half of the concepts appear in fewer than ten images, whereas a small number of highly generic concepts account for a substantial fraction of the annotations (Figure~\ref{fig:label-distribution}). This severe class imbalance represents one of the central challenges of the task. The dataset also exhibits considerable visual heterogeneity. Images differ in their state of preservation, digitization quality, cropping strategy, and layout, with some folios containing multiple independent scenes. Such variability provides a realistic benchmark for evaluating the robustness and generalization capabilities of modern computer vision models.


\section{Methodological Framework}

This work is designed as a comparative evaluation of two complementary paradigms for automatic iconographic annotation.

\paragraph{Vision-Language Models.} The first paradigm relies on VLMs. We adapt a recent Qwen model through lightweight LoRA fine-tuning so that it directly generates annotations constrained to the vocabulary of the TIMEL thesaurus. This approach benefits from the broad visual and linguistic knowledge acquired during large-scale multimodal pre-training and offers considerable flexibility for handling rare or poorly represented concepts. However, it does not explicitly exploit the hierarchical organization of the thesaurus and remains sensitive to prompt design as well as to the generation of unsupported predictions (hallucinations).

\paragraph{Supervised discriminative classifiers (SDC).}
The second paradigm formulates iconographic annotation as supervised extreme multi-label classification over the fixed TIMEL vocabulary. All configurations share a DINOv3 visual encoder and compare three complementary classification heads: a flat linear classifier operating on the global CLS representation, the taxonomy-aware CAT decoder, which traverses the TIMEL hierarchy level by level, and ML-Decoder, which learns to aggregate spatial evidence directly from the patch tokens without imposing an explicit taxonomic structure. Unlike generative VLMs, these classifiers assign scores exclusively to predefined TIMEL concepts and therefore cannot produce out-of-vocabulary labels. 

\paragraph{Evaluation.} A particular emphasis is placed on the evaluation protocol. Conventional multi-label classification metrics treat all prediction errors equally and therefore ignore the semantic proximity between related concepts belonging to the same branch of the taxonomy. We consequently introduce evaluation measures that explicitly account for hierarchical proximity within the thesaurus, allowing us to distinguish completely incorrect predictions from semantically related ones. Such an evaluation is essential in the context of computer-assisted annotation for cultural heritage collections.

\paragraph{Research questions.} Beyond comparing overall predictive performance, this study addresses three research questions:

\begin{itemize}
    \item RQ1. How do Vision-Language Models and supervised discriminative classifiers (SDC) differ in their ability to support automated iconographic annotation of medieval illuminations?
    \item RQ2. Does explicit hierarchical decoding improve large-scale iconographic retrieval compared with flat multi-label classification?
    \item RQ3. How does the descriptive or interpretative nature of iconographic concepts affect model performance?
\end{itemize}

The last question constitutes one of the central hypotheses of this work. We hypothesize that different model families excel at different levels of semantic abstraction: perceptual recognition for descriptive concepts versus contextual reasoning for interpretative iconographic concepts.


\section{Methods}

This section will present the two architectures evaluated in this work:


\subsection{Evaluation Framework}

We evaluate every model as a multi-label predictor: for each image the prediction $P$ and the ground
truth $G$ are \emph{sets} of canonical TIMEL leaves. Since TIMEL is a taxonomy, we complement
exact-match scoring with hierarchy-aware measures. We write $\mathrm{depth}(c)$ for the depth of a
node (root $=0$), $\mathrm{anc}(c)$ for its ancestors including $c$ and the root, and
$\mathrm{LCA}(a,b)$ for the lowest common ancestor; the five TIMEL roots are disjoint, so two nodes
in different roots share no ancestor.

\paragraph{Flat metrics.} Treating $P$ and $G$ as sets, we report micro- and macro-averaged
precision, recall and $F_1$, plus the exact-match rate $\mathbf{1}[P=G]$. Micro scores aggregate
$\mathrm{TP}=\sum_e|P_e\cap G_e|$, $\mathrm{FP}=\sum_e|P_e\setminus G_e|$ and
$\mathrm{FN}=\sum_e|G_e\setminus P_e|$ over all examples $e$; macro scores average the per-class
$F_1$ over the classes present in the gold. Under an assistive-annotation reading, where the model
returns a \emph{ranked} list of suggestions (for our VLM the JSON emission order, for supervised
baselines the class-probability order), we additionally report $\mathrm{precision}@1$,
$\mathrm{recall}@K$ and $F_1@K$ for $K\in\{5,10,20\}$, and the mean reciprocal rank
$\mathrm{MRR}=\frac{1}{|\mathcal{E}|}\sum_{e}1/\mathrm{rank}_e$ where $\mathrm{rank}_e$ is the
position of the first correct leaf. Here $\mathrm{precision}@K=|\{\text{top-}K\}\cap G|/K$ and
$\mathrm{recall}@K=|\{\text{top-}K\}\cap G|/|G|$, and $F_1@K$ is their harmonic mean. $F_1@K$ is the
headline ranked score; $\mathrm{precision}@1$ (a strict ``is the top suggestion correct'') and
$\mathrm{recall}@K$ (coverage of a $K$-item shortlist) are reported for completeness, while
$\mathrm{MRR}$ grades how early the first correct leaf appears.

\paragraph{Hierarchical metrics.} To credit predictions that land in the right region of the
taxonomy, we use two complementary families. The \emph{hierarchical} micro $F_1$
(Kiritchenko--Verspoor) compares ancestor sets $A(P)=\bigcup_{p\in P}\mathrm{anc}(p)$ and
$A(G)=\bigcup_{g\in G}\mathrm{anc}(g)$, aggregating $\mathrm{TP}=\sum_e|A_e(P)\cap A_e(G)|$ (and
$\mathrm{FP}$, $\mathrm{FN}$ likewise) across examples. The \emph{soft} $F_1$ instead scores each
pair through
\[
\mathrm{sim}_{\mathrm{soft}}(a,b)=\frac{\mathrm{depth}(\mathrm{LCA}(a,b))}{\max(\mathrm{depth}(a),\mathrm{depth}(b))},
\]
aggregated as
$P_{\mathrm{soft}}=\frac{1}{|P|}\sum_{p\in P}\max_{g\in G}\mathrm{sim}_{\mathrm{soft}}(p,g)$ with a
symmetric recall $R_{\mathrm{soft}}$ (roles of $P$ and $G$ swapped), both macro-averaged over
examples. The two differ in how they treat a shared root: a prediction and a gold sharing
\emph{only} the root contribute a positive true positive to the hierarchical score (the root lies in
both ancestor sets) but score $0$ under $\mathrm{sim}_{\mathrm{soft}}$. As TIMEL is heavily
long-tailed, we also report an IC-weighted soft $F_1$ that weights each leaf by its information
content $\mathrm{IC}(c)=-\log\!\big((\mathrm{count}(c)+\alpha)/(N+\alpha|V|)\big)$ (additive
smoothing $\alpha=1$, $|V|=2\,333$), so that errors on rare concepts count for more.

\paragraph{Output validity (VLM-specific).} A generative model can emit strings outside the
controlled vocabulary, so we report the \emph{invalid rate} (hallucination)
$\mathrm{invalid}=n_{\mathrm{invalid}}/(n_{\mathrm{valid}}+n_{\mathrm{invalid}})$ over all emitted
labels. A supervised classifier restricted to the label set scores $0$ by construction, whereas a
VLM can be non-zero. Appendix~\ref{app:metrics-example} works all of these metrics through on a
single TIMEL instance.


\subsection{Vision Language Model (VLM)}

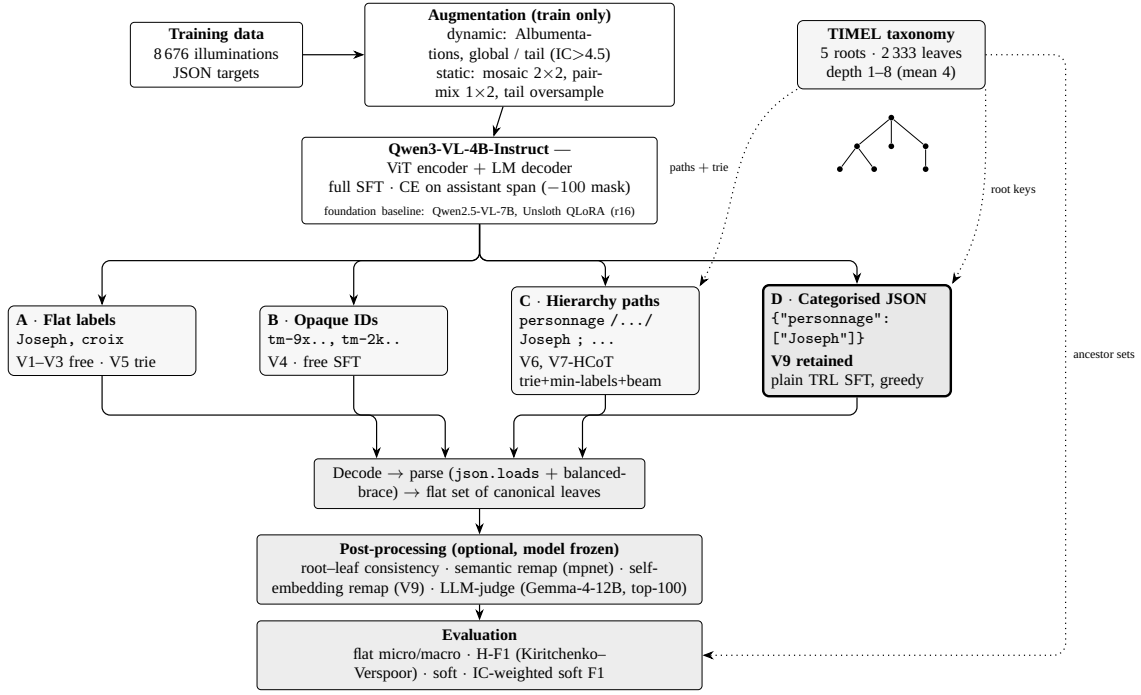
\begin{figure*}[t]
\centering
\resizebox{\textwidth}{!}{%
\begin{tikzpicture}[
  font=\scriptsize, >=Stealth,
  box/.style ={draw, rounded corners=2pt, align=center, inner sep=3pt},
  fam/.style ={draw, rounded corners=3pt, align=left,   inner sep=4pt, text width=30mm, fill=black!3},
  stg/.style ={draw, rounded corners=2pt, align=center, inner sep=4pt, fill=black!7},
  res/.style ={draw, rounded corners=3pt, align=center, inner sep=4pt, text width=30mm, fill=black!4},
  dot/.style ={circle, fill=black, inner sep=1pt},
  arr/.style ={->, semithick},
  con/.style ={->, dotted, semithick},
]

\node[box, text width=28mm] (train) at (2.0, 5.0)
  {\textbf{Training data}\\8\,676 illuminations\\JSON targets};
\node[box, text width=52mm] (aug) at (7.3, 5.0)
  {\textbf{Augmentation (train only)}\\dynamic: Albumentations, global / tail (IC$>$4.5)\\static: mosaic 2$\times$2, pairmix 1$\times$2, tail oversample};
\draw[arr] (train) -- (aug);

\node[res] (tax) at (13.8, 5.0)
  {\textbf{TIMEL taxonomy}\\5 roots $\cdot$ 2\,333 leaves\\depth 1--8 (mean 4)};

\node[box, text width=60mm] (bb) at (6.6, 2.8)
  {\textbf{Qwen3-VL-4B-Instruct} --- ViT encoder $+$ LM decoder\\full SFT $\cdot$ CE on assistant span ($-100$ mask)\\[2pt]{\tiny foundation baseline: Qwen2.5-VL-7B, Unsloth QLoRA (r16)}};
\draw[arr] (aug) -- (bb);

\begin{scope}[shift={(13.8, 3.4)}]
  \node[dot](r0) at (0, 0.5) {};
  \node[dot](r1) at (-0.6, 0) {};
  \node[dot](r2) at (0, 0) {};
  \node[dot](r3) at (0.6, 0) {};
  \node[dot](r4) at (-0.9, -0.4) {};
  \node[dot](r5) at (-0.3, -0.4) {};
  \node[dot](r6) at (0.6, -0.4) {};
  \draw (r0)--(r1) (r0)--(r2) (r0)--(r3) (r1)--(r4) (r1)--(r5) (r3)--(r6);
\end{scope}

\node[fam] (A) at (0, 0)
  {\textbf{A $\cdot$ Flat labels}\\\texttt{Joseph, croix}\\[2pt]V1--V3 free $\cdot$ V5 trie};
\node[fam] (B) at (4.4, 0)
  {\textbf{B $\cdot$ Opaque IDs}\\\texttt{tm-9x.., tm-2k..}\\[2pt]V4 $\cdot$ free SFT};
\node[fam] (C) at (8.8, 0)
  {\textbf{C $\cdot$ Hierarchy paths}\\\texttt{personnage /.../}\\\texttt{Joseph ; ...}\\[2pt]V6, V7-HCoT\\trie+min-labels+beam};
\node[fam, fill=black!9, very thick] (D) at (13.2, 0)
  {\textbf{D $\cdot$ Categorised JSON}\\\texttt{\{"personnage":}\\\texttt{["Joseph"]\}}\\[2pt]\textbf{V9 retained}\\plain TRL SFT, greedy};

\draw[arr, rounded corners=4pt] (bb.south) -- (6.6, 1.4) -| (A.north);
\draw[arr, rounded corners=4pt] (bb.south) -- (6.6, 1.4) -| (B.north);
\draw[arr, rounded corners=4pt] (bb.south) -- (6.6, 1.4) -| (C.north);
\draw[arr, rounded corners=4pt] (bb.south) -- (6.6, 1.4) -| (D.north);

\draw[con] (tax.south west) to[out=210, in=60] node[above left,font=\tiny,yshift=-1mm]{paths $+$ trie} (C.north east);
\draw[con] (tax.south east) to[out=270, in=60] node[right,font=\tiny]{root keys} (D.north east);

\node[stg, text width=55mm] (dec) at (6.6, -2.5)
  {Decode $\rightarrow$ parse (\texttt{json.loads} $+$ balanced-brace) $\rightarrow$ flat set of canonical leaves};

\draw[arr, rounded corners=4pt] (A.south) -- (0, -1.3) -| ([xshift=-18mm]dec.north);
\draw[arr, rounded corners=4pt] (B.south) -- (4.4, -1.3) -| ([xshift=-6mm]dec.north);
\draw[arr, rounded corners=4pt] (C.south) -- (8.8, -1.3) -| ([xshift=6mm]dec.north);
\draw[arr, rounded corners=4pt] (D.south) -- (13.2, -1.3) -| ([xshift=18mm]dec.north);

\node[stg, text width=75mm] (pp) at (6.6, -4.0)
  {\textbf{Post-processing (optional, model frozen)}\\ root--leaf consistency $\cdot$ semantic remap (mpnet) $\cdot$ self-embedding remap (V9) $\cdot$ LLM-judge (Gemma-4-12B, top-100)};
\draw[arr] (dec) -- (pp);

\node[stg, text width=75mm] (eval) at (6.6, -5.5)
  {\textbf{Evaluation}\\ flat micro/macro $\cdot$ H-F1 (Kiritchenko--Verspoor) $\cdot$ soft $\cdot$ IC-weighted soft F1};
\draw[arr] (pp) -- (eval);

\draw[con, rounded corners=6pt] (tax.east) -- ++(1.4, 0) |- node[right,font=\tiny,pos=0.25]{ancestor sets} (eval.east);

\end{tikzpicture}%
}
\caption{VLM pipeline for TIMEL classification. All strategies share one backbone
(\texttt{Qwen3-VL-4B-Instruct}, full SFT; a QLoRA \texttt{Qwen2.5-VL-7B} baseline reported
separately) and diverge only at three stages. \emph{Training-side augmentation} (dynamic
Albumentations and static mosaic/pairmix/oversampling, both long-tail-oriented) feeds the SFT.
Four \emph{output-encoding families} differ in target representation and decoding constraint:
(A) flat French labels, optionally trie-constrained (V5); (B) opaque \texttt{tm-xxxx} identifiers
(V4, abandoned); (C) explicit root-to-leaf paths with trie, minimum-label and beam constraints
(V6/V6.1 and the staged HCoT variant V7); (D) the retained categorised-JSON target (V9), a bare
TRL \texttt{SFTTrainer} decoded greedily. Predictions are parsed to a flat leaf set, optionally
refined by \emph{post-processing} (consistency, semantic/self-embedding remap, multimodal
LLM-judge), then scored. The TIMEL taxonomy supplies trie prefixes (A-V5, C), JSON category keys
and the consistency filter (D), and the ancestor sets used by the hierarchical metrics.}
\label{fig:vlm}
\end{figure*}

We frame iconographic indexing as multi-label classification of medieval illuminations against the
TIMEL controlled vocabulary: $2\,333$ annotatable leaf concepts drawn from a $4\,660$-node taxonomy
with five roots (\emph{nature lieu}, \emph{objet architecture}, \emph{personnage}, \emph{sujet},
\emph{th\`eme}) and depth $1$--$8$ (mean $4$); each image bears $5$--$35$ labels (mean ${\approx}10$).
Every model is fine-tuned on the same split from one
backbone, \texttt{Qwen3-VL-4B-Instruct}, so that strategies differ only in how the label set is
represented in the output, how decoding is constrained, and how the training set is
augmented---never in the visual encoder. Supervision is next-token cross-entropy on the assistant
span only ($-100$ mask elsewhere); an early defect that located this span by matching the raw string
\texttt{<|im\_start|>assistant} (unstable under Qwen's BPE tokeniser) was corrected by matching the
token-ID sequence. Shared optimiser settings are learning rate $2\!\times\!10^{-5}$, cosine
schedule, \texttt{bf16} with gradient checkpointing, and a vision budget of
$[256,512]\!\times\!28\!\times\!28$ pixels (upper bound $401\,408$); warm-up and weight decay differ
by generation ($0.03$ / $0.05$ for the legacy full-SFT runs, $0.1$ / $0$ for the retained V9). A
parameter-efficient foundation baseline is also reported on \texttt{Qwen2.5-VL-7B} with Unsloth
QLoRA (4-bit; rank $16$, $\alpha\,16$, dropout $0$; adapters on all linear layers plus
\texttt{lm\_head}/\texttt{embed\_tokens}; lr $2\!\times\!10^{-4}$). The strategies fall into four
output-encoding families, a training-side augmentation stage, and an inference-side post-processing
stage (Figure~\ref{fig:vlm}).

\paragraph{(A) Natural-language flat labels.} The target is the comma-separated list of canonical
French labels, generated under one fixed user instruction and an explored system prompt. V1 is plain
SFT with greedy decoding; a defect that masked the assistant span by matching the raw string
\texttt{<|im\_start|>assistant} was corrected in V2 by matching the token-ID sequence, together with a
systematic sweep of system prompts and generation parameters. V3 keeps plain SFT but prepends a
structured prefix to the flat list (a hybrid target), still without any decode constraint. V5 leaves
both the target and the SFT objective unchanged and instead constrains \emph{inference}: a
\texttt{LabelTrieLogitsProcessor} built over the $2\,333$ leaf strings masks, at each step, every
token that does not continue a valid label prefix, with two trie variants per leaf (with and without
a leading space) to absorb the BPE leading-space effect. The trie removes paraphrase but the model
still collapses onto frequent classes and rarely emits the long tail.

\paragraph{(B) Opaque identifiers.} V4 keeps plain SFT and greedy decoding but swaps the surface
labels for the vocabulary's \texttt{tm-xxxx} identifiers, testing whether removing natural language
suppresses paraphrase. It is abandoned: the model learns the identifier \emph{form} (a fixed prefix
plus eight characters) but never binds an identifier to an image, which isolates grounding---not
paraphrase---as the primary obstacle.

\paragraph{(C) Explicit hierarchy.} V6 rewrites every target on the fly inside the collator
(\texttt{rewrite\_messages\_to\_paths}, from the flat split plus the taxonomy, with no offline
preprocessing), serialising each leaf as its full root-to-leaf path (\texttt{\ /\ } within a path,
\texttt{\ ;\ } between paths) so that shared intermediate nodes supervise rare leaves
compositionally; the trie is rebuilt over the $2\,333$ complete paths, the training schedule is
lengthened, and \texttt{max\_new\_tokens} is raised to $1\,280$ for the longer sequences
(\texttt{no\_repeat\_ngram\_size}$=5$, \texttt{repetition\_penalty}$=1.1$). Because the model tends to
stop early, V6.1 stacks training- and decode-time patches: class-balanced resampling (with
replacement, each example weighted by the maximum information content of its labels,
\texttt{balance\_factor}$=2.0$) to over-represent rare classes; depth-sorted targets to stabilise the
emission order over shared prefixes; a trie filtered to annotatable leaves; a
\texttt{MinLabelsLogitsProcessor} that blocks EOS until \texttt{min\_labels}$=5$ separators are
emitted; and beam search (\texttt{num\_beams}$=4$, \texttt{length\_penalty}$=1.2$). The trie and the
minimum-label mask can jointly forbid every token---yielding a \texttt{nan} softmax and a
deadlock---so the minimum-label constraint applies a $-30$ attenuation rather than $-\infty$. V7
restructures the target into a staged chain-of-thought built by \texttt{compute\_hcot\_target}---
\texttt{[BRANCHES]} (the present roots), \texttt{[REFINE]} (their depth-$2$ ancestors) and
\texttt{[LEAVES]} (the leaves, depth-sorted)---and decodes \emph{section-aware}: the trie and
minimum-label constraints activate only after the \texttt{[LEAVES]} marker, leaving the reasoning
prefix free. The custom-tag format proved less reliable than JSON for a modern VLM.

\paragraph{(D) Categorised JSON (retained).} V9 keeps the sole surviving differentiator---injecting
the taxonomy into the output format---and drops all decoding machinery. The target is a JSON object
grouping labels under the TIMEL roots actually present,
\texttt{\{"categories":\{"personnage":[\dots],\dots\}\}}, which forces a commitment to branches
before leaves (the V7 intuition, in a format VLMs see abundantly during pre-training). Training is a
bare TRL \texttt{SFTTrainer} with greedy inference: no custom collator, no logits processor, no trie,
no minimum-label forcing, no class balancing. Robustness is moved from decoding to data
handling---images cast for lazy PIL decoding, \texttt{min\_pixels}/\texttt{max\_pixels} passed to the
processor, \texttt{max\_length}$=$\texttt{None} to defeat \texttt{SFTConfig}'s default $1\,024$-token
truncation (which would clip image tokens), and the per-part \texttt{image} path stripped to avoid an
Arrow schema unification that injects spurious \texttt{<|image\_pad|>} tokens; the generated string is
read back with \texttt{json.loads} and a balanced-brace fallback. A representative run uses batch
$2\times$grad-accum $32$ (effective $64$) for one epoch (${\approx}136$ steps). The retained prompt
(system message and one-line user instruction, identical at training and inference) is reproduced
verbatim in Figure~\ref{fig:prompt}. Formally, the fine-tuned model $p_\theta$ emits this JSON string
autoregressively and the prediction is
$\hat{Y}_{\mathrm{V9}}=\mathrm{leaves}\big(\mathrm{parse}(\arg\max_{y}\,p_\theta(y\mid x,\pi))\big)$
under greedy decoding, where $x$ is the image and $\pi$ the fixed prompt.

\begin{figure*}[t]
\centering
\begin{minipage}{0.92\textwidth}
\rule{\textwidth}{0.4pt}\par\vspace{2pt}
\footnotesize\textsc{System prompt}\par\vspace{-4pt}
{\footnotesize
\begin{verbatim}
Tu es un classifieur visuel iconographique TIMEL. Pour chaque image d'enluminure
médiévale, réponds STRICTEMENT au format JSON suivant :

{"categories": {"<racine TIMEL>": ["<label>", "<label>", ...], ...}}

Les racines TIMEL sont exactement : "nature lieu", "objet architecture",
"personnage", "sujet", "thème". Tu n'utilises que les racines qui sont
effectivement représentées dans l'image. Chaque label doit être un libellé
canonique du référentiel TIMEL (orthographe exacte en français), pas une
paraphrase. Pas de doublon. Termine après la fermeture du JSON.
\end{verbatim}}
\vspace{-4pt}\footnotesize\textsc{User turn} (with one image token)\par\vspace{-4pt}
{\footnotesize
\begin{verbatim}
Générer les mots-clés timel pour cette image.
\end{verbatim}}
\vspace{-6pt}\rule{\textwidth}{0.4pt}
\end{minipage}
\caption{Retained V9 prompt, The system message
fixes the categorised-JSON schema and the five TIMEL roots, forbids paraphrases and duplicates, and
requires canonical French labels; the user turn is a single fixed instruction paired with one image
token per example.}
\label{fig:prompt}
\end{figure*}

\paragraph{Training-side augmentation.} Because macro-averaged and IC-weighted scores are dominated
by rare leaves, two augmentation regimes target the long tail. \emph{Dynamic} augmentation is applied stochastically per access through
an Albumentations pipeline---\texttt{ColorJitter}/\texttt{RGBShift} ($p{=}0.6$), \texttt{Perspective}
($p{=}0.3$), \texttt{GaussianBlur} ($p{=}0.2$) and an aggressive \texttt{CoarseDropout} ($4$--$10$
holes covering $5$--$15\%$, $p{=}0.6$)---either on every example (\texttt{global}) or only on
examples carrying a rare label of information content $\mathrm{IC}>4.5$ (\texttt{tail}); validation
is never augmented. \emph{Static} augmentation instead materialises images on disk and concatenates
them to the training set: a $2\times2$ \texttt{mosaic} and a $1\times2$ \texttt{pairmix} of whole
illuminations ($448$-px tiles) whose label is the per-root union, plus \texttt{oversample} variants
of tail examples (horizontal flip $p{=}0.5$, rotation $\pm5^\circ$ $p{=}0.6$, light perspective
$p{=}0.4$, random erasing, and blur-or-JPEG degradation $p{=}0.5$; defaults $2\,000$ mosaics,
$3\,000$ pairmixes, oversampling factor $2$). Colour jitter is deliberately excluded from
oversampling---illumination colours are semantically meaningful---and CutMix was removed because,
absent object masks, a pasted patch cannot be guaranteed to carry the motif justifying its label.

\paragraph{Inference-side post-processing.} A set of optional, chainable steps refines the JSON
prediction without touching the model. A \emph{root--leaf consistency} filter drops any leaf whose
canonical root in the taxonomy disagrees with the JSON category it was placed under. Two
\emph{semantic-remap} steps rescue out-of-vocabulary paraphrases by embedding each invalid term and
attaching its nearest canonical neighbour above a cosine threshold of $0.7$---either with a
multilingual sentence encoder (\texttt{paraphrase-multilingual-mpnet-base-v2}) or with the
fine-tuned V9 model's own mean-pooled last-layer hidden states (its $2\,333$ canonical embeddings
cached). Finally, a multimodal \emph{LLM-as-judge} (Gemma-4-12B, run in-process, optionally 4-bit)
receives the image, the V9 prediction and a root-balanced top-$100$ candidate list, and may add,
remove or correct labels within that set. An orchestrator chains these steps (\texttt{consistency},
\texttt{semantic}, \texttt{semantic\_vlm}, \texttt{judge}), passing each output to the next and
retaining intermediates for ablation.

\paragraph{Zero-shot baseline and mixture-of-experts.} In parallel to
fine-tuning, we test whether a strong general VLM reaches competitive predictions \emph{without any
training}, using only prompting and deterministic post-calibration; the backbone here is
\texttt{Gemini-3-Flash}, queried through an API. Three configurations of increasing structure are
compared. The \emph{zero-shot} baseline issues a single call, mapping the image $x$ and a prompt
listing the controlled vocabulary to a parsed leaf set $\hat{Y}_0=\mathrm{leaves}(g(x))$, and serves
as the reference for whether task-specific adaptation is needed at all. Injecting the full
controlled vocabulary into the prompt follows prior work on LLM annotators with non-standard
tagsets, where it acts as an explicit in-context constraint against out-of-schema
labels~\cite{vidal-gorene-etal-2026-resourced}. The
\emph{mixture-of-experts} (MoE) configuration decomposes the taxonomy by root: for each of the five
roots $r$, an expert $E$ reads the image and only the indented subtree $T_r$ ($330$ to $650$ labels
instead of $2\,333$) and returns a deliberately generous candidate set $E(x,T_r)$. The union passes
through a fuzzy out-of-vocabulary recovery $\phi$ (nearest canonical label, difflib cutoff $0.86$)
and then to a taxonomic judge $J$, which sees each candidate with its path and siblings and may keep
it, drop it, or replace it by a sibling:
\[
C=\phi\!\Big(\textstyle\bigcup_{r\in R}E(x,T_r)\Big),\qquad
\hat{Y}_{\mathrm{MoE}}=J\big(x,\{(c,\mathrm{path}(c),\mathrm{sib}(c)):c\in C\}\big).
\]
Each stage targets one metric: the root experts raise recall, $\phi$ suppresses hallucination, and
the judge restores precision and granularity.

\paragraph{Verify-calibrate reranking (V9 $+$ MoE).} A final stage reranks the \emph{pooled}
candidates of the fine-tuned V9 model and the MoE experts (their union over-predicts, mean
cardinality $28$ against a gold mean of ${\approx}11.5$), cutting false positives through scoring and
deterministic calibration without any new generation. Writing $C$ for this pooled candidate set, a
scorer grades every candidate from the image on two integer axes, visual evidence
$v(c)\in\{0,\dots,10\}$ and annotation propensity $u(c)\in\{0,\dots,10\}$ (whether a TIMEL annotator
would index it, since body parts, ordinary garments and ubiquitous decoration are visible yet
conventionally not indexed), giving a scalar
\[
s(c)=\frac{v(c)\,u(c)}{100}\in[0,1].
\]
Two calibration steps use train statistics only. A \emph{propensity cap} bounds, over the $N$ test
images, the occurrences of a label $\ell$ by $\mathrm{cap}(\ell)=\max\!\big(2,\lceil 2\,p(\ell)\,N
\rceil\big)$, with $p(\ell)$ its train document-frequency, dropping the lowest-scored excess (the
floor of $2$ protects rare labels). A \emph{cardinality-calibrated threshold} $\tau$ then keeps the
$\lceil \bar{c}\,N\rceil$ highest-scored image--label pairs, with $\bar{c}=11.9$ the mean train
cardinality and no access to test gold; each image emits its surviving labels in decreasing $s$, so
the emission order is a genuine confidence ranking. This transposes to a generative VLM the
ingredients that make the discriminative classifiers strong, a ranked output, a calibrated cut and
frequency priors, without any gradient training.

\begin{figure*}[ht!]
\centering
{\setlength{\fboxsep}{1.5pt}%
\newcommand{\gchip}[1]{\colorbox[HTML]{d6e6fb}{\textcolor[HTML]{1b4a99}{\strut #1}}}%
\newcommand{\tpchip}[1]{\colorbox[HTML]{d6f5e0}{\textcolor[HTML]{0c5e2c}{\strut #1}}}%
\newcommand{\fnchip}[1]{\colorbox[HTML]{fde5e3}{\textcolor[HTML]{952420}{\strut #1}}}%
\newcommand{\dchip}[1]{\colorbox[HTML]{ececec}{\textcolor[HTML]{555555}{\strut #1}}}%
\begin{minipage}[t]{0.31\textwidth}
\vspace{0pt}
\centering
\includegraphics[width=\linewidth]{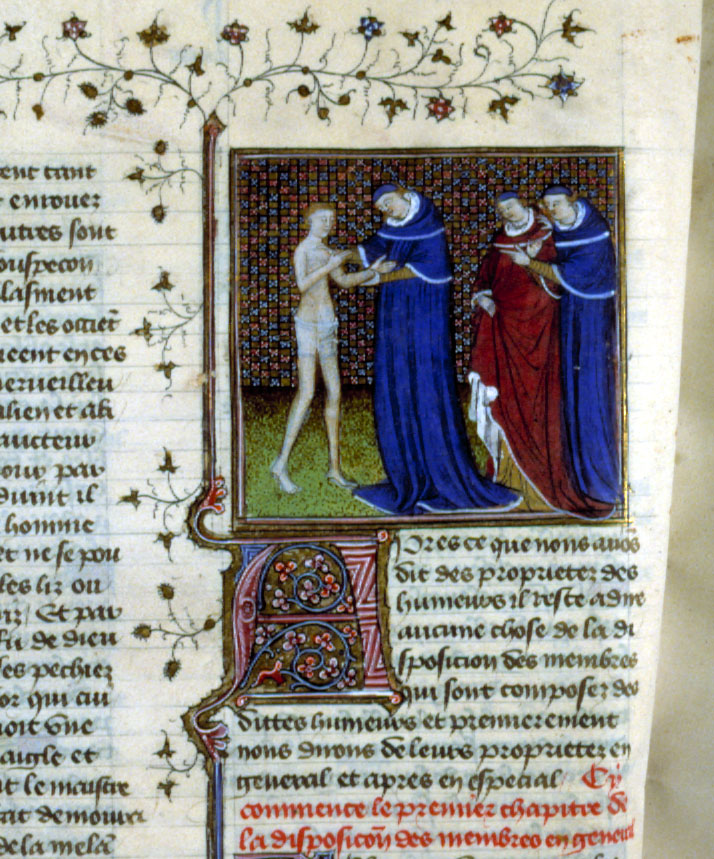}\\[3pt]
{\footnotesize Folio \texttt{003333}: a nude patient examined by robed physicians (a medical scene).}\par\vspace{3pt}
{\scriptsize\raggedright \textbf{Gold (7):}\\ \gchip{scène de médecine}\, \gchip{médecine}\, \gchip{médecin}\, \gchip{malade}\, \gchip{anatomie}\, \gchip{nudité}\, \gchip{scène d'astronomie}\par}
\end{minipage}\hfill
\begin{minipage}[t]{0.65\textwidth}
\vspace{0pt}
{\footnotesize\textbf{Verify--calibrate reranking (V9 $+$ MoE).} The pooled candidates of the
fine-tuned V9 model and the MoE experts ($49$ labels) are scored $s=v\cdot u/100$ from the image,
capped by the frequency prior ($15$ removed), and cut by the calibrated cardinality threshold,
leaving $4$ labels. Here precision is $4/4$, recall $4/7$, with no invalid term.}\par\vspace{4pt}
\scriptsize
\begin{tabular}{clccp{0.42\linewidth}}
\toprule
rank & label & $v$ & $u$ & evidence (verifier) \\
\midrule
\multicolumn{5}{l}{\textbf{Retained} (ranked by $s$; all true positives)}\\
1 & \tpchip{scène de médecine} & 10 & 10 & auscultation of a patient by practitioners \\
2 & \tpchip{médecine} & 9 & 10 & central theme of medical knowledge \\
3 & \tpchip{médecin} & 9 & 10 & robed figure performing the examination \\
4 & \tpchip{malade} & 9 & 10 & passive, undressed subject being examined \\
\midrule
\multicolumn{5}{l}{\textbf{Gold missed}}\\
-- & \fnchip{nudité} & 8 & 8 & bare torso and legs, just below the threshold \\
-- & \fnchip{anatomie} & 7 & 8 & body inspected by clerics, below the threshold \\
-- & \fnchip{scène d'astronomie} & -- & -- & not proposed by V9 or MoE (recall ceiling) \\
\bottomrule
\end{tabular}\par\vspace{4pt}
{\scriptsize\raggedright
\textbf{Dropped, high-score over-readings:}\, \dchip{clerc}\, \dchip{enseignement}\, \dchip{scène d'enseignement}\, \dchip{toucher}\, \dchip{moine}\, \dchip{peau}\, \dchip{maladie}\par\vspace{2pt}
\textbf{Capped by frequency prior (visible but not indexed):}\, \dchip{désignation}\, \dchip{discussion}\, \dchip{témoin}\, \dchip{livre}\, \dchip{écrit}\, \dchip{parole}\, \dchip{plante}\, \dchip{fleur}\, \dchip{cheveux}\, \dchip{oreille}\, \dchip{saisie}\, \dchip{présentation}\, \dchip{spectateur}\, \dchip{disciple}\, \dchip{intérieur}\ (+ low-score body-part/generic labels \emph{main, tête, bras, jambe, pied, œil, vêtement, tissu, parchemin, pavement, feuille}).\par}
\end{minipage}%
}
\caption{Qualitative example of verify--calibrate reranking on folio \texttt{003333}. The fine-tuned
V9 model and the zero-shot MoE experts jointly propose $49$ candidate labels; the verifier scores each
from the image on visual evidence $v$ and annotation propensity $u$ ($s=v\cdot u/100$), the propensity
cap removes over-frequent ``visible but not indexed'' labels (body parts, decoration, generic
actions), and the calibrated threshold keeps only the highest-scored labels. The output is a compact,
fully correct set (precision $4/4$); the gold labels \emph{nudité} and \emph{anatomie} are scored but
fall just below the threshold, while \emph{scène d'astronomie} is never proposed (a recall ceiling
inherited from the candidate stage). Colours: \colorbox[HTML]{d6e6fb}{\textcolor[HTML]{1b4a99}{gold}},
\colorbox[HTML]{d6f5e0}{\textcolor[HTML]{0c5e2c}{retained (TP)}},
\colorbox[HTML]{fde5e3}{\textcolor[HTML]{952420}{missed (FN)}},
\colorbox[HTML]{ececec}{\textcolor[HTML]{555555}{dropped}}.}
\label{fig:example-003333}
\end{figure*}

\paragraph{Experimental Results}

Every model is scored as a multi-label predictor with the metrics of the evaluation framework, after
projecting its output to a flat set of canonical TIMEL leaves. The ranking used by
$\mathrm{prec@1}$, MRR and $F_1@K$ is the emission order for the generative models and the score
order $s(c)$ for verify-calibrate; $\mathrm{precision}@K=|\{\text{top-}K\}\cap G|/K$, consistent with
the discriminative protocol (top-$K$, no decision threshold). Table~\ref{tab:vlm-results} reports the
VLM family, spanning the nine fine-tuning versions (output-encoding families~A--D), the zero-shot
baselines, and the mixture-of-experts with reranking.

\begin{table}[t]
\centering
\caption{VLM results across the fine-tuning strategies (families~A/C/D), the zero-shot baselines and
the training-free pipeline. Ranked metrics ($\mathrm{prec@1}$, MRR, $F_1@K$) use the emission order
(score order for verify-calibrate); soft-$F_1$ and hier-$F_1$ follow the evaluation framework.
\emph{card} is the mean predicted cardinality (gold mean ${\approx}11$) and \emph{inv.} the
invalid-output rate. Bold row = best retained configuration per group.}
\label{tab:vlm-results}
\resizebox{\textwidth}{!}{%
\begin{tabular}{lcccccccc}
\toprule
System
  & prec@1$\uparrow$ & MRR$\uparrow$
  & $F_1$@10$\uparrow$ & $F_1$@20$\uparrow$
  & soft-$F_1$$\uparrow$ & hier-$F_1$$\uparrow$
  & card & inv.\%$\downarrow$ \\
\midrule
\multicolumn{9}{l}{\textit{(1)~Fine-tuned SFT (Qwen3-VL-4B), versions by output-encoding family}} \\
V1 flat labels (A)                  & 0.224 & 0.429 & 0.230 & 0.175 & 0.366 & 0.437 & 7.4  & 0.0 \\
V2 flat, span/prompt fix (A)        & 0.298 & 0.472 & 0.236 & 0.183 & 0.398 & 0.451 & 8.2  & 3.4 \\
V3 flat, structured prefix (A)      & 0.337 & 0.518 & 0.241 & 0.194 & 0.421 & 0.463 & 8.9  & 2.1 \\
V4 opaque IDs (B)                   & 0.209 & 0.408 & 0.184 & 0.134 & 0.348 & 0.407 & 6.6  & 7.6 \\
V5 flat, trie decoding (A)          & 0.361 & 0.539 & 0.238 & 0.186 & 0.409 & 0.452 & 7.1  & 0.0 \\
V6 hierarchy paths (C)              & 0.650 & 0.700 & 0.118 & 0.075 & 0.310 & 0.262 & 2.3  & 29.2 \\
V6.1 paths + min-labels + beam (C)  & 0.500 & 0.613 & 0.121 & 0.081 & 0.264 & 0.345 & 6.7  & 5.0 \\
V7 HCoT staged paths (C)            & 0.412 & 0.566 & 0.206 & 0.164 & 0.431 & 0.451 & 8.6  & 6.1 \\
\textbf{V9 categorised JSON (D)}    & \textbf{0.420} & \textbf{0.575} & \textbf{0.247} & \textbf{0.209} & \textbf{0.460} & \textbf{0.486} & 10.4 & 4.5 \\
\midrule
\multicolumn{9}{l}{\textit{(2)~Zero-shot (reference, no training)}} \\
Gemini-3-Flash              & 0.211 & 0.404 & 0.258 & 0.330 & 0.507 & 0.514 & 22.4 & 4.3 \\
Qwen2.5-VL-72B              & 0.316 & 0.482 & 0.190 & 0.138 & 0.334 & 0.300 & 23.0 & 18.3 \\
\midrule
\multicolumn{9}{l}{\textit{(3)~Mixture-of-experts and reranking}} \\
MoE (5 experts + judge)         & 0.421 & 0.540 & 0.255 & 0.319 & 0.482 & 0.504 & 28.0 & 0.0 \\
VC top-11 (ablation)            & 0.526 & 0.675 & 0.364 & 0.255 & 0.481 & 0.487 & 11.0 & 0.0 \\
VC no-prior (ablation)          & 0.526 & 0.661 & 0.336 & 0.283 & 0.547 & 0.547 & 11.9 & 0.0 \\
\textbf{VC rerank (V9 $+$ MoE)} & \textbf{0.526} & \textbf{0.661} & \textbf{0.337} & \textbf{0.287} & \textbf{0.561} & \textbf{0.559} & 11.9 & 0.0 \\
\bottomrule
\end{tabular}%
}
\end{table}

Within the fine-tuning family, the progression across output-encoding strategies is what matters. The
natural-language flat target improves steadily from V1 to V3 as the assistant-span defect is fixed
and the prompt is tuned; the opaque-identifier target (V4) brings no advantage over labels, and trie
decoding (V5) trades a little coverage for zero hallucination. The path-based runs (V6, V6.1) then
regress \emph{below} the flat baseline on coverage: forcing explicit root-to-leaf paths makes the
model stop early and under-predict (V6 emits only $2.3$ labels), which inflates prec@1 to $0.650$
while collapsing $F_1$@10 to $0.118$ and driving hallucination to $29\%$; the staged HCoT variant
(V7) recovers most of this. The categorised-JSON target (V9) is the strongest configuration on every
coverage and hierarchy metric (hier-$F_1$ $0.486$, soft-$F_1$ $0.460$, $F_1$@10 $0.247$), indicating
that the gain comes from the output format itself rather than from constrained decoding.

The zero-shot models fix the reference point. Fine-tuning helps where expected: the retained V9
(Qwen3-VL-4B) places a correct concept at rank~1 on $42\%$ of images (MRR $0.575$), above both
zero-shot baselines ($0.211$/$0.404$ and $0.316$/$0.482$ for prec@1/MRR), while emitting a compact
set (card $10.4$) with few invalid terms ($4.5\%$). The larger \texttt{Qwen2.5-VL-72B} is the weakest
system and hallucinates most ($18.3\%$), so raw scale without adaptation does not suffice. The
generous zero-shot \texttt{Gemini-3-Flash} (card $22.4$) instead trades precision for coverage and
already attains high hierarchical scores (hier-$F_1$ $0.514$), showing that a strong general VLM
captures much of the taxonomy unsupervised.

Adding structure closes the gap further. The mixture-of-experts raises recall
(recall@$20$ $0.53$) at the cost of precision (card $28$); reranking the pooled V9 and MoE candidates
(verify-calibrate) then recovers precision and is the best VLM configuration overall, leading on
prec@1 ($0.526$), MRR
($0.661$), $F_1$@10 ($0.337$), soft-$F_1$ ($0.561$) and hier-$F_1$ ($0.559$) at a gold-matched
cardinality ($11.9$) with zero invalid output. The propensity cap adds a small but consistent gain
over the threshold alone, whereas a fixed top-$11$ is clearly worse, the gold cardinality ranging
from $1$ to $38$. In line with RQ3, this gain concentrates on descriptive concepts (branch $F_1$
rising to $0.500$) and stays marginal on interpretative ones ($0.397$), whose residual false
positives are plausible scene readings that attribute-level scoring does not remove.

Relative to the discriminative classifiers (Table~\ref{tab:disc-results}), the best VLM
(verify-calibrate) exceeds the strongest head (ML-Decoder\,+\,CLS) on hierarchical $F_1$ ($0.559$
versus $0.496$) but remains below it on the ranked scores (prec@1 $0.526$ versus $0.696$, MRR $0.661$
versus $0.806$, $F_1$@10 $0.337$ versus $0.380$) and on soft-$F_1$ ($0.561$ versus $0.638$), on their
respective test subsets. Figure~\ref{fig:example-003333} traces the reranking on a representative
folio.


\subsection{Supervised discriminative classifiers (SDC)}

Following the Vision-Language Model experiments, we turn to supervised discriminative classifiers to
investigate the architectural choices for this extreme multi-label classification (XMC) task. We
compare three classification heads sharing the same frozen visual backbone (DINOv3), enabling
controlled ablations of hierarchical modelling, spatial pooling, backbone adaptation, and loss
function.

The \emph{flat linear classifier} projects the CLS token (a dedicated token prepended to the patch
sequence that, after passing through all self-attention layers, aggregates global image content into
a single vector $\mathbf{z}_\mathrm{CLS}\in\mathbb{R}^{1024}$) onto $C$ independent binary outputs,
serving as the primary data-level and loss baseline.
\emph{Hierarchical cross-attention transformer} (CAT)~\cite{springstein2024} replaces the flat
output with a cross-attention decoder that traverses the TIMEL taxonomy level by level, enforcing
taxonomic consistency at inference time. Finally, \emph{ML-Decoder}~\cite{ridnik2023} bypasses the
CLS representation altogether by attending directly to the 196 spatial patch tokens extracted by
DINOv3 through 100 learned group queries, without imposing any explicit assumption on the label
hierarchy.

The four hypotheses H1~(loss function), H2~(backbone adaptation via LoRA), H3~(spatial pooling
vs.\ global CLS) and H4~(descriptive vs.\ interpretative concepts) are tested through controlled
ablations within and across architectures.
Table~\ref{tab:disc-results} reports the full ablation experiments.

\paragraph{Common visual encoder: DINOv3 ViT-L/16.}
All three classifiers share the same frozen visual backbone, \textbf{DINOv3 ViT-L/16}
(\texttt{facebook/dinov3-vitl16-pretrain-lvd1689m}), a self-supervised Vision Transformer with
300M parameters distilled from a 7B teacher pre-trained on LVD-1689M (1.689~billion curated web images)~\cite{dinov3}. The choice of DINOv3 over alternative visual encoders rests on three properties specific to our setting.
First, self-supervised training produces domain-agnostic representations free from ImageNet label
biases and from the language-alignment constraints of CLIP-style models, a desirable property when
transferring to medieval illuminations whose visual categories overlap poorly with natural image
benchmarks. Second, unlike CLIP-based encoders used in prior iconographic work~\cite{springstein2024},
DINOv3 preserves dense spatial structure through its patch tokens: the grid of $P=196$ spatial
representations $\mathbf{H}\in\mathbb{R}^{B\times196\times1024}$ retains locality, which is
critical for recognising concepts that occupy only a small region of a folio (attributes, tools,
background figures). Third, the architecture natively exposes two complementary representations: a
global summary token $\mathbf{z}_\mathrm{CLS}\in\mathbb{R}^{B\times1024}$ and the full spatial
sequence; whose respective utility can be contrasted in a controlled experimental design. The
backbone is frozen in all configurations except in LoRA experimentations.

\paragraph{Experimental strategies and hypotheses.}
Four independent hypotheses structure the experimental sequence.

\emph{H1. Loss function.}
With $C=2{,}384$ classes and two to eight positive labels per image, over 99\,\% of gradient mass
under standard binary cross-entropy (BCE) comes from easy negatives (see Appendix~\ref{app:losses}
for formal definitions). Asymmetric Loss (ASL)~\cite{ridnik2021} decouples positive and negative
focusing factors ($\gamma^+{=}0$, $\gamma^-{=}4$) and applies a probability margin $m{=}0.05$,
strongly downweighting easy negatives without attenuating positive gradients.
We hypothesise that this loss substitution alone will substantially improve retrieval without any
architectural change.

\emph{H2. Backbone adaptation via LoRA.}
We hypothesise that low-rank adaptation~\cite{hu2022} of the attention projections
(\texttt{q\_proj} and \texttt{v\_proj}, rank $r$) can further improve retrieval, acknowledging the
risk of over-adaptation on a corpus of limited size.

\emph{H2-bis. LoRA on all linear projections.}
H2 targets only the query and value projections (\texttt{q\_proj}, \texttt{v\_proj}). H2-bis asks
whether this scope restriction explains the observed degradation. All six linear projections of
DINOv3 ViT-L are targeted: attention (\texttt{q\_proj}, \texttt{k\_proj}, \texttt{v\_proj},
\texttt{o\_proj}) and MLP (\texttt{up\_proj}, \texttt{down\_proj}). Rank is reduced to $r=8$
(vs.\ $r=16$ in H2) to partially offset the larger adapter count; \texttt{backbone\_lr\_multiplier}
$=0.1$ is kept. If H2-bis improves over H2, the degradation is attributable to insufficient scope;
if it still degrades, the frozen DINOv3 representations are inherently more robust than any
low-rank adaptation at this corpus scale, regardless of which projections are adapted.

\emph{H3. Spatial pooling vs.\ global CLS.}
We hypothesise that explicitly attending over the 196~patch tokens will yield richer multi-label
predictions than projecting a single global summary, particularly for spatially localised concepts.

\emph{H3-bis. CLS token as additional context for ML-Decoder.}
H3 contrasts two \emph{architectures}: CLS projection vs.\ patch-level cross-attention. H3-bis
asks a finer question within the ML-Decoder family: does prepending the CLS token to the
196~patch tokens, yielding a 197-token key-value context, improve predictions beyond patch tokens
alone? The CLS token aggregates a global image summary after all 24 self-attention layers of
DINOv3; it may supply scene-level context that complements the local spatial evidence in the patch
tokens, particularly for concepts that require viewing the folio as a whole. All other
settings are identical to the best ML-Decoder baseline; only the encoder output is extended from
$\mathbf{H}\in\mathbb{R}^{B\times196\times D}$ to
$[\mathbf{z}_\mathrm{CLS};\,\mathbf{H}]\in\mathbb{R}^{B\times197\times D}$.

\emph{H4. Descriptive-vocabulary specialisation.}
Building on the distinction between descriptive and interpretative concepts introduced in
Section~\ref{sec:dataset}, we test whether restricting supervision to the descriptive
vocabulary improves the retrieval of visually grounded concepts. We re-train the
best-performing architecture (ML-Decoder + CLS + aug., selected from
Table~\ref{tab:disc-results}) using only the \emph{nature lieu} and
\emph{objet architecture} branches. Images without descriptive annotations are excluded,
leaving 9,460 records and 769 active leaf classes. Training, validation, and evaluation
operate on this filtered vocabulary, and the resulting model is compared with the same
architecture trained on the full vocabulary but evaluated on the corresponding descriptive
subset. We hypothesise that descriptive-only training will improve retrieval by removing
competition from interpretative labels. This experiment directly addresses RQ3 and is
reported in Table~\ref{tab:branch-ablation}.

\emph{Architecture choices.} The three architectures represent complementary strategies for extreme
multi-label classification and a long-tail distribution of the TIMEL thesaurus. The flat linear
classifier serves as the reference baseline, CAT explicitly leverages the hierarchical structure of
TIMEL, whereas ML-Decoder learns to aggregate spatial evidence directly from DINOv3 patch tokens.
Their comparison enables controlled evaluation of the four hypotheses introduced above.

\paragraph{(1) Flat linear baseline.}
The simplest head projects the CLS token onto $C$ independent binary logits
$\hat{\mathbf{y}}=\sigma(W\mathbf{z}_\mathrm{CLS}+\mathbf{b})$, $W\in\mathbb{R}^{C\times1024}$,
directly testing the linear separability of TIMEL concepts in the DINOv3 CLS space. It is
evaluated under BCE and ASL~(H1), and with LoRA backbone adaptation~(H2, rank $r=16$). The
no-augmentation run (Flat-BCE, no aug., Table~\ref{tab:disc-results}) serves as the data-level
reference; all other flat runs and all subsequent architectures use the shared augmentation pipeline
described below.

\paragraph{(2) CAT: hierarchical cross-attention decoder.}
CAT~\cite{springstein2024} replaces the flat output with a Transformer decoder that traverses the
TIMEL taxonomy one level at a time. At each depth~$\ell$, a cross-attention sublayer takes the
current decoder state as query and the 196~patch tokens as key-value context, and produces a
distribution over the children of the currently active node:
$\mathbf{z}^{(\ell)}=\operatorname{CrossAttn}(\mathbf{q}^{(\ell)},\mathbf{H},\mathbf{H})$,
$p(c_\ell\mid c_{<\ell},x)=\operatorname{softmax}(W^{(\ell)}\mathbf{z}^{(\ell)})$.
Training minimises a hierarchical cross-entropy (hier-CE) accumulated over all active tree levels
(Appendix~\ref{app:losses}), providing supervision at intermediate nodes and benefiting rare leaves
that share parents with more frequent siblings. A \emph{beam search} of width~$K$ maintains the
$K$ most probable partial paths at each level. Every predicted leaf is taxonomically consistent with
its ancestors by construction. However, an incorrect decision at an upper level eliminates all
correct descendants in that sub-tree (this coverage risk  grows when gold labels span several
disjoint TIMEL branches). The hier-ASL ablation applies asymmetric loss at each level of the
traversal, addressing the same easy-negative domination that persists within each node's local child
distribution.

\paragraph{(3) ML-Decoder: learned spatial pooling.}
ML-Decoder~\cite{ridnik2023} retains the full patch-token sequence and introduces $G=100$ learned
group queries attending over it:
$\mathbf{U}=\operatorname{MultiHeadAttn}(\mathbf{Q},\mathbf{H},\mathbf{H})$,
$\mathbf{Q}\in\mathbb{R}^{100\times1024}$.
Each query output is mapped to a subset of $C$ classes via a GroupFC layer, pooling spatially
localised evidence. Unlike CAT, ML-Decoder encodes no assumption about the taxonomy structure: the
100~queries do not correspond to TIMEL branches, and all classes are scored independently. The head
is trained with ASL, directly testing H3. A further variant prepends the CLS token to the patch
sequence, extending the cross-attention context to 197~tokens (H3-bis).

\paragraph{Training environment.}
\textit{Shared settings.} PyTorch (2.8.0), Lightning (2.5.5), and Weights \& Biases (0.28.1) were
used to implement, train, and monitor the proposed models. All runs are performed on one NVIDIA RTX
PRO 1000 Blackwell GPU (8~GiB VRAM) with bf16 mixed precision. All experiments use AdamW
($\text{lr}=10^{-4}$, weight decay $10^{-4}$), linear warm-up (3~epochs) followed by cosine
annealing, input resolution $224\!\times\!224$, and a maximum of 50~epochs with early stopping on
validation loss (patience~20). The training split contains approximately 9,689~images; the 46
evaluation images are excluded from training, validation, and checkpoint selection throughout.

\textit{Data augmentation.}
A preliminary flat-BCE run without data augmentation serves as the data-level reference; all
subsequent experiments apply the same shared pipeline: random resized crop (scale $[0.7,\,1.0]$),
horizontal flip ($p{=}0.5$), ColorJitter (brightness and contrast $\pm0.2$, saturation $\pm0.1$,
$p{=}0.5$), and GaussianBlur (kernel $3$--$7$, $p{=}0.2$).

\textit{LoRA settings.}
H2 and H2-bis are evaluated on the flat-linear head only. Adapters target \texttt{q\_proj} and
\texttt{v\_proj} (H2, $r{=}16$) or all six linear projections \texttt{q}, \texttt{k}, \texttt{v},
\texttt{o\_proj} and MLP \texttt{up}/\texttt{down\_proj} (H2-bis, $r{=}8$).
LoRA was not applied to the CAT or ML-Decoder heads: the flat-head evidence in H2 and H2-bis was judged sufficient to conclude
that backbone perturbation does not improve retrieval at this corpus scale.

\paragraph{Results.}
Predictions are obtained by ranking logits and retaining the top-$K$ entries ($K\in\{10,20\}$),
without a per-model decision threshold. Results are computed on the same test dataset used for
the VLM experiments.

\begin{table}[t]
\centering
\caption{Full results for discriminative classifiers experiments on 46~fixed images (seed~42). Predictions
are ranked by logit score; the top $N{=}20$ entries are retained without a sigmoid threshold.
Bold row = best retained configuration per architecture group. Loss formulas:
Appendix~\ref{app:losses}.}
\label{tab:disc-results}
\resizebox{\textwidth}{!}{%
\begin{tabular}{lllcccccc}
\toprule
Configuration & Loss & Backbone
  & prec@1$\uparrow$ & MRR$\uparrow$
  & $F_1$@10$\uparrow$ & $F_1$@20$\uparrow$
  & soft-$F_1$$\uparrow$ & hier-$F_1$$\uparrow$ \\
\midrule
\multicolumn{9}{l}{\textit{(1)~Flat linear classifier}} \\
Flat-BCE (no aug.)                              & BCE      & frozen         & 0.435 & 0.588 & 0.238 & 0.235 & 0.528 & 0.384 \\
Flat-BCE + aug.                                 & BCE      & frozen         & 0.435 & 0.589 & 0.246 & 0.239 & 0.530 & 0.387 \\
Flat-BCE + aug.\ + LoRA ($r{=}16$) [H2]         & BCE      & LoRA $r{=}16$  & 0.413 & 0.584 & 0.234 & 0.213 & 0.511 & 0.367 \\
\textbf{Flat-ASL + aug.\ [H1]}                  & ASL      & frozen         & 0.565 & 0.697 & 0.304 & 0.302 & 0.596 & 0.444 \\
Flat-ASL + aug.\ + LoRA ($r{=}16$) [H2]         & ASL      & LoRA $r{=}16$  & 0.500 & 0.612 & 0.195 & 0.194 & 0.496 & 0.348 \\
Flat-ASL + aug.\ + LoRA-all ($r{=}8$) [H2-bis]  & ASL      & LoRA-all $r{=}8$ & 0.522 & 0.647 & 0.236 & 0.223 & 0.523 & 0.373 \\
\midrule
\multicolumn{9}{l}{\textit{(2)~CAT: cross-attention hierarchical decoder}} \\
CAT-beam + aug.                                  & hier-CE  & frozen         & 0.565 & 0.678 & 0.231 & 0.212 & 0.345 & 0.285 \\
CAT-beam + aug.\ + hier-ASL [H1]                  & hier-ASL & frozen         & 0.609 & 0.701 & 0.248 & 0.231 & 0.551 & 0.359 \\
\midrule
\multicolumn{9}{l}{\textit{(3)~ML-Decoder: learned spatial pooling}} \\
ML-Decoder + aug.\ (BCE)                         & BCE      & frozen         & 0.696 & 0.764 & 0.364 & 0.349 & 0.635 & 0.493 \\
ML-Decoder + aug.\ [H3]                          & ASL      & frozen         & 0.696 & 0.801 & 0.376 & 0.338 & 0.629 & 0.485 \\
\textbf{ML-Decoder + CLS + aug.\ [H3-bis]}       & ASL      & frozen         & \textbf{0.696} & \textbf{0.806} & \textbf{0.380} & \textbf{0.348} & \textbf{0.638} & \textbf{0.496} \\
\bottomrule
\end{tabular}%
}
\end{table}

\begin{table}[t]
\centering
\caption{H4 branch-level evaluation. Both rows use ML-Decoder + CLS + aug.\ (best from
Table~\ref{tab:disc-results}); the H4 model is re-trained exclusively on the \emph{descriptive}
vocabulary (\emph{nature lieu}, \emph{objet architecture}). Metrics are filtered to descriptive
labels in both rows (top $N{=}20$, no sigmoid threshold). Bold row = best value per column.}
\label{tab:branch-ablation}
\resizebox{\textwidth}{!}{%
\begin{tabular}{llcccccc}
\toprule
Architecture & Training vocabulary & prec@1$\uparrow$ & MRR$\uparrow$ & $F_1$@10$\uparrow$ & soft-$F_1$$\uparrow$ & hier-$F_1$$\uparrow$ \\
\midrule
ML-Decoder + CLS + aug.\ [H3-bis]             & full vocabulary  & 0.609 & 0.719 & \textbf{0.342} & \textbf{0.638} & \textbf{0.513} \\
ML-Decoder + CLS + aug.\ (H4 re-trained) [H4] & descriptive only & \textbf{0.674} & \textbf{0.752} & 0.331 & 0.607 & 0.402 \\
\bottomrule
\end{tabular}%
}
\end{table}

\emph{H1 is confirmed for the flat-linear head and nuanced for ML-Decoder.}
On the CLS-linear architecture, switching from BCE to ASL yields
$\Delta\,\text{prec@1}{=}{+}0.130$, $\Delta\,\text{MRR}{=}{+}0.108$, and
$\Delta\,F_1\text{@10}{=}{+}0.058$, the largest single gain across all ablations.
For the ML-Decoder head, the loss effect is substantially weaker and metric-dependent. BCE and ASL
achieve virtually identical prec@1 ($0.696$ vs.\ $0.696$), while ASL leads on MRR ($0.801$
vs.\ $0.764$, $\Delta{=}{+}0.037$) and $F_1$@10 ($0.376$ vs.\ $0.364$, $\Delta{=}{+}0.012$).
Conversely, BCE slightly outperforms ASL on coverage and hierarchical metrics ($F_1$@20: $0.349$
vs.\ $0.338$; soft-$F_1$: $0.635$ vs.\ $0.629$; hier-$F_1$: $0.493$ vs.\ $0.485$). A plausible
explanation is that the ML-Decoder's 100~group queries already provide an implicit specialisation
mechanism that redistributes capacity across the class space, reducing the class-imbalance
sensitivity that makes ASL beneficial for the flat head. The overall message is that loss
calibration is a prerequisite for the flat classification head.

\emph{H2 is not confirmed.}
Adapting the backbone with LoRA ($r{=}16$, backbone lr $10^{-5}$) on the flat head under BCE
yields $\Delta\,\text{prec@1}{=}{-}0.022$ relative to the frozen baseline, suggesting that
DINOv3's CLS representations are already sufficiently transferable at this data scale and that
backbone perturbation introduces more noise than useful signal. Switching
to ASL does not rescue LoRA adaptation on the flat head: Flat-ASL + LoRA ($r{=}16$) degrades on
every metric relative to the frozen Flat-ASL baseline ($\Delta\,\text{prec@1}{=}{-}0.065$,
$\Delta\,\text{MRR}{=}{-}0.085$, $\Delta\,F_1\text{@10}{=}{-}0.109$). The uniform deterioration
across precision, recall, and hierarchical metrics rules out a simple precision/recall trade-off and
points instead to representation collapse.

\emph{H2-bis is not confirmed.} Expanding LoRA to all six projections at reduced rank ($r{=}8$)
partially recovers the H2 degradation
($\Delta\,\text{prec@1}{=}{+}0.022$, $\Delta\,\text{MRR}{=}{+}0.035$,
$\Delta\,F_1\text{@10}{=}{+}0.041$ vs.\ Flat-ASL + LoRA $r{=}16$) but remains below the frozen
baseline on every metric
($\Delta\,\text{prec@1}{=}{-}0.043$, $\Delta\,\text{MRR}{=}{-}0.050$,
$\Delta\,F_1\text{@10}{=}{-}0.068$, $\Delta\,\text{soft-}F_1{=}{-}0.073$,
$\Delta\,\text{hier-}F_1{=}{-}0.071$). The persistent gap confirms that the degradation observed
in H2 is not attributable to an insufficient scope of adaptation: backbone perturbation via LoRA
consistently hurts the flat-linear head regardless of which projections are targeted, pointing to a
structural incompatibility between low-rank adaptation and the CLS-based pooling regime at this
corpus scale. Given these results, LoRA fine-tuning of the CAT and ML-Decoder heads was not pursued: the H2 and H2-bis evidence on the flat head provides a sufficient basis for concluding
that backbone perturbation does not improve retrieval at this data scale, independently of the
classification head.

\emph{H3 is confirmed.} ML-Decoder + aug.\ outperforms the best CLS-projection alternative
(Flat-ASL + aug.) by $\Delta\,\text{prec@1}{=}{+}0.131$, $\Delta\,\text{MRR}{=}{+}0.104$,
$\Delta F_1\text{@10}{=}{+}0.072$, $\Delta\,\text{soft-}F_1{=}{+}0.033$, and
$\Delta\,\text{hier-}F_1{=}{+}0.041$, confirming that the 196~patch tokens suppressed in the CLS
summary carry discriminative spatial information for TIMEL concept recognition.

\emph{H3-bis is not conclusive.} Prepending the CLS token to the 196~patch tokens yields small but
directionally consistent improvements across all metrics ($\Delta\,\text{MRR}{=}{+}0.005$,
$\Delta\,F_1\text{@10}{=}{+}0.004$, $\Delta\,\text{soft-}F_1{=}{+}0.009$,
$\Delta\,\text{hier-}F_1{=}{+}0.011$). However, on a 46-image evaluation subset, a single
correct or incorrect prediction shifts $F_1$@10 by approximately ${\pm}0.02$; The observed gains are therefore too small relative to the granularity of the evaluation to be reliably distinguished from random variation.

\emph{H4 is partially confirmed.} Table~\ref{tab:branch-ablation} compares ML-Decoder + CLS
trained on the full vocabulary (evaluated on the descriptive subset only) against the same
architecture re-trained exclusively on descriptive labels (9,460 records, 769 active leaves).
Re-training on the descriptive vocabulary improves ranking-oriented metrics: prec@1 rises
($\Delta\text{prec@1}{=}{+}0.065$) and MRR increases similarly
($\Delta\text{MRR}{=}{+}0.033$), indicating that the specialised model more reliably places
correct descriptive concepts near the top of the ranked list. Coverage-oriented and hierarchical
metrics, however, move in the opposite direction: $F_1$@10 decreases slightly
($\Delta\text{$F_1$}{=}{-}0.011$), soft-$F_1$ drops moderately (
$\Delta\text{soft-$F_1$}{=}{-}0.031$), and hierarchical $F_1$ drops substantially (
$\Delta\text{hierarchical $F_1$}{=}{-}0.111$). A plausible explanation is that interpretative labels act as implicit context
during full-vocabulary training: descriptive and interpretative concepts co-occur frequently within
the same image (a scene annotated with \emph{annonciation à la Vierge} necessarily also carries \emph{ange} and
architectural labels), so the full-vocabulary model acquires richer co-occurrence signals that
improve hierarchical recall of descriptive concepts. The restricted H4 model, by contrast, trains
without this cross-branch contextual signal, achieving sharper top-$K$ ranking precision at the
cost of broader hierarchical coverage. These results suggest that focussing training on the
descriptive vocabulary sharpens the ranking head, a property directly useful in a annotation workflow. Another structural explanation: with only 769 active classes, ML-Decoder may be over-parameterised for the reduced descriptive task, limiting coverage gains despite improved top-K ranking.

\paragraph{Synthesis.}
The experiments identify loss calibration and spatial feature aggregation as the main determinants
of retrieval quality. ASL substantially improves the flat-linear classifier by mitigating
easy-negative domination, while ML-Decoder achieves the strongest overall performance by attending
directly to DINOv3 patch tokens. Although CAT benefits from hierarchical ASL, its branch-wise
decoding strategy still limits coverage across disjoint regions of the taxonomy.

Neither LoRA adaptation nor descriptive-only training yields consistent improvements. LoRA
systematically degrades retrieval, suggesting that frozen DINOv3 representations are already
sufficiently transferable at this corpus scale. Restricting training to descriptive labels improves
top-ranked precision but reduces hierarchical coverage, indicating that cross-branch co-occurrences
provide useful supervision. Overall, the main bottleneck appears to lie in spatial feature
aggregation rather than backbone adaptation or explicit taxonomy modelling. Taxonomic knowledge may
therefore be more effectively incorporated through the loss function, training-data organisation,
or post-processing than directly enforced by the decoder architecture.

At the operational level, ML-Decoder + CLS + aug.\ places a correct TIMEL concept at rank~1 in
roughly 70\% of cases (prec@1${}=0.696$) and recovers labels from the interpretative branches
(\emph{personnage}, \emph{sujet}, \emph{thème}) that correspond to gold annotations.
A recurring limitation is the over-prediction of corpus co-occurrents. In battle or duel scenes, for example, the model frequently assigns high scores to several weapon-related concepts from the same taxonomic branch, such as \emph{épée}, \emph{bouclier}, \emph{arc} and \emph{haubert}, even when only one or two are visually present or annotated. This behaviour suggests that the model relies partly on corpus-level co-occurrence priors rather than exclusively on concept-specific visual evidence. The uniformly high scores assigned to these associated labels may also indicate overconfidence or insufficient probability calibration, inflating the predicted set and reducing precision at larger $K$. Preliminary inspection suggests that a sigmoid threshold in the $[0.65,0.70]$ range could suppress part of this co-occurrence-driven noise while retaining the correct labels; systematic confidence calibration is left for future work.


\section{Conclusion}

\paragraph{Limitations.} The results should first be interpreted in light of several limitations. The annotated corpus remains relatively small, and the evaluation was conducted on only 46 images. The long-tail distribution of concepts, the possible incompleteness of the annotations, and the absence of probabilistic calibration call for caution when interpreting some performance differences. Moreover, the comparison between VLMs and supervised classifiers should be considered carefully, as the former rely on multi-stage generation and calibration procedures, whereas the latter perform direct discriminative prediction. Consequently, the conclusions reported here require validation on a larger and more diverse corpus, including future efforts to expand and refine the available annotations.

\paragraph{RQ1.} The first research question reveals less a difference in overall capability than a difference in the nature of the knowledge exploited by the models and the documentary tasks for which they are best suited. VLMs benefit from large-scale pre-training, which provides semantic priors extending beyond the visual regularities observed in the supervised corpus (as shown by the Gemini-3-Flash zero-shot baseline). This broad prior knowledge allows them to generate plausible concepts beyond those directly represented in the training annotations. However, experiments V1-V9 show that their main challenge lies not in proposing concepts, but in aligning them with a controlled vocabulary: the improvement obtained with the categorized JSON representation (V9) indicates that an appropriate output structure is critical for reducing this semantic gap. By contrast, supervised discriminative classifiers (SDCs) learn a direct mapping between visual representations and TIMEL labels. Trained on the annotated corpus, they reproduce the indexing conventions encoded in the controlled vocabulary, making them particularly effective for retrieval within a closed concept space. This specialization also exposes them to corpus-specific regularities: experiment H4 suggests that they can exploit frequent concept co-occurrences, which may sometimes introduce predictions supported by annotation patterns rather than explicit visual evidence alone.

\paragraph{RQ2.} The second research question shows that explicitly incorporating the TIMEL hierarchy provides a useful documentary constraint, but does not necessarily yield the best multi-label prediction strategy. Compared with ML-Decoder, CAT guarantees taxonomic consistency of predicted paths; however, an error at an upper level may redirect the search toward an incorrect branch and prevent retrieval of relevant concepts located in other subtrees. This observation is consistent with the VLM experiments, where structured outputs (V9) improve performance more effectively than complex hierarchical decoding strategies. Overall, the thesaurus hierarchy appears more valuable as structured knowledge to exploit than as a decision space to traverse. To support this evaluation, we deliver the soft F1 metric as a contribution for taxonomy-aware scoring, providing a finer assessment of hierarchical errors based on lowest common ancestors.

\paragraph{RQ3.} Finally, the third research question confirms that model behaviour strongly depends on the nature of the concepts being predicted. Descriptive concepts particularly benefit from discriminative approaches: H4 shows that retraining the best classifier on the descriptive vocabulary alone improves the ranking of visually grounded concepts, although at the expense of hierarchical coverage. This suggests that joint learning with interpretative concepts provides useful documentary context through cross-branch correlations. Conversely, interpretative concepts represent a domain where VLMs may provide complementary semantic hypotheses through their broader pre-training, although aligning these hypotheses with controlled documentary categories remains challenging. The two approaches therefore address different needs: supervised classifiers provide greater consistency with established indexing practices, whereas VLMs facilitate semantic exploration and hypothesis generation.

\paragraph{Hybrid architecture.} Taken together, these findings suggest that VLMs and SDCs should not be viewed as competing approaches, but as complementary components of future documentary indexing systems. VLMs provide semantic exploration and generalization capabilities, while SDCs deliver consistent predictions within a closed documentary space. These results motivate the development of hybrid multimodal architectures combining large-scale pre-trained visual representations, supervised specialization on TIMEL, and explicit exploitation of thesaurus knowledge. Such architectures could further integrate additional sources of information available at the folio level—including text transcriptions, descriptive metadata, and external knowledge from digital libraries—to improve both the quality and contextualization of automatic indexing.

\paragraph{Deployment sustainability.} Beyond predictive performance, practical deployment requires considering inference complexity and scalability. SDCs provide a direct and efficient mapping to the TIMEL vocabulary once trained, whereas VLM-based approaches may involve more complex pipelines combining generation, output structuring, candidate expansion, and reranking (e.g., MoE and Verify-Calibrate). VLMs are therefore particularly suited to exploratory workflows, annotation assistance, and hypothesis generation, while supervised classifiers offer a practical and lightweight solution for systematic large-scale indexing. Rather than replacing expert practices, future heritage indexing systems should combine these complementary capabilities according to the documentary task.








\newpage


\FloatBarrier
\appendix
\section{Annexe}


\subsection{Corpus statistics}
\label{app:corpus-stats}

\begin{figure}[H]
    \centering
    \includegraphics[width=0.9\linewidth]{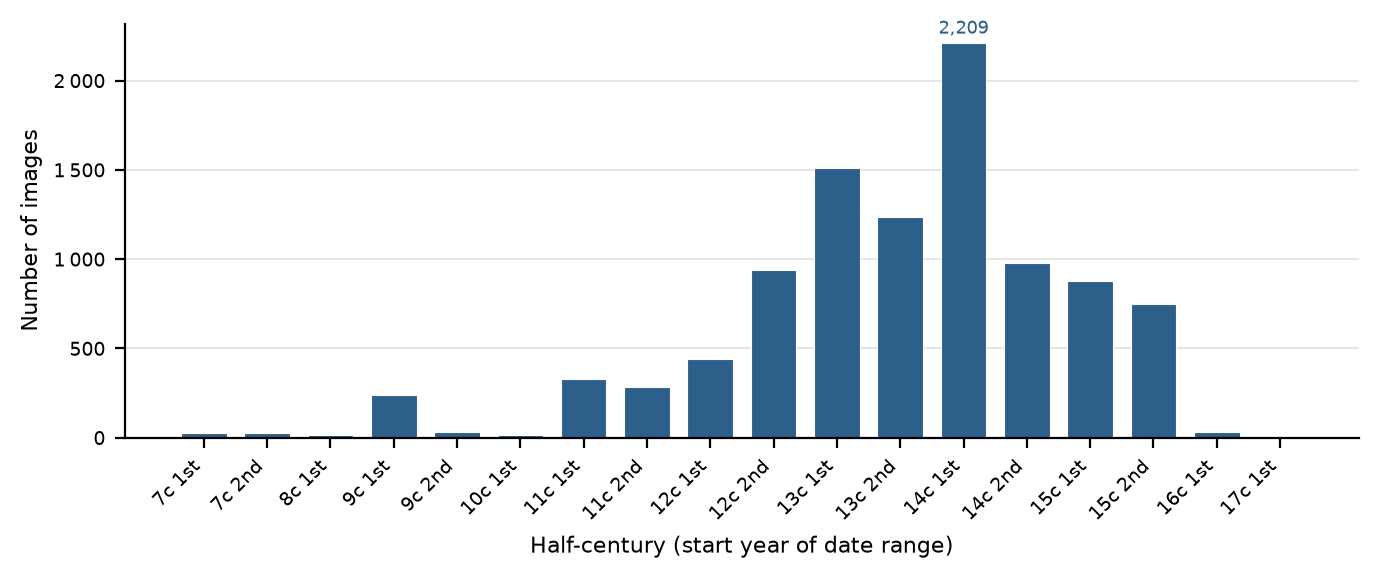}
    \caption{Temporal distribution of the 9,735 folios.}
    \label{fig:corpus-dates}
\end{figure}

\begin{figure}[H]
    \centering
    \includegraphics[width=0.9\linewidth]{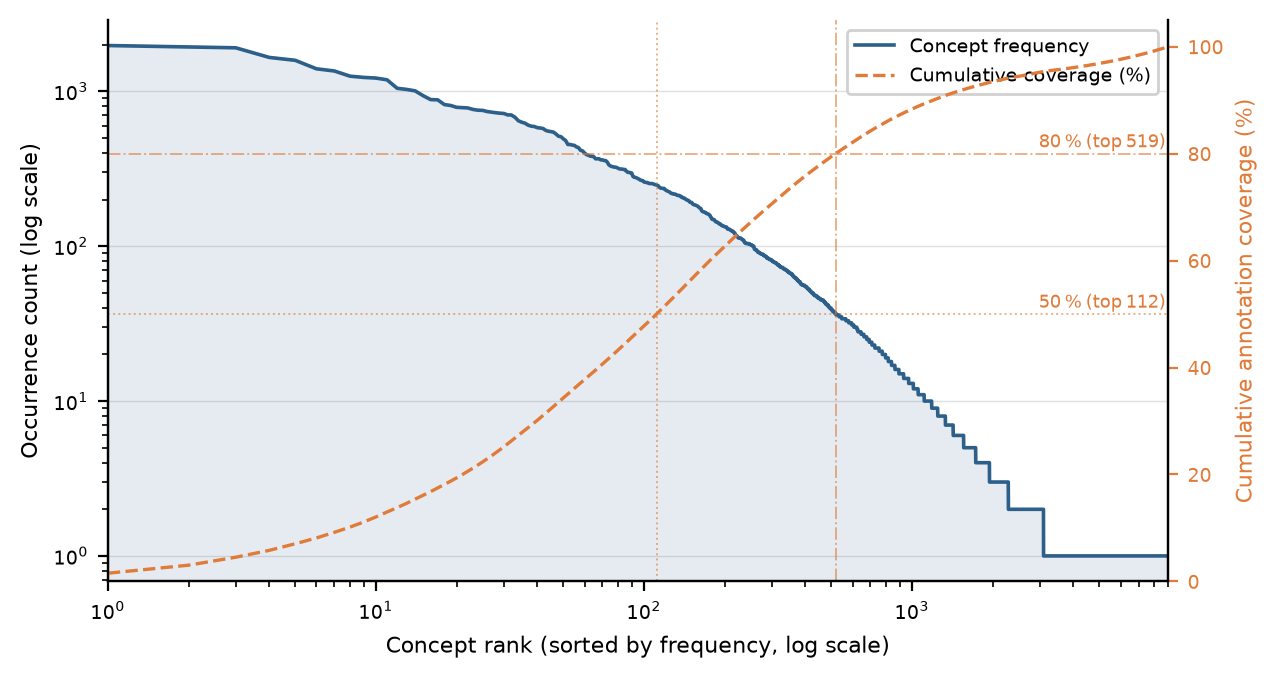}
    \caption{Long-tail distribution of TIMEL concept frequencies (log-log scale). Concepts are ranked by
    frequency; the near-linear slope confirms a Zipfian law, with frequency dropping below 100
    occurrences by rank~100 and most of the 9,024 concepts appearing only once. The dashed curve
    (right orange axis) shows cumulative annotation coverage: the top 112 concepts account for 50\,\% and
    the top 519 for 80\,\% of all annotations.}
    \label{fig:label-distribution}
\end{figure}
\FloatBarrier

\subsection{Worked example: evaluation metrics on a TIMEL instance}
\label{app:metrics-example}

\begin{figure}[t]
\centering
\resizebox{\textwidth}{!}{%
\begin{tikzpicture}[
  font=\scriptsize,>=Stealth,
  nd/.style ={draw,rounded corners,inner sep=2pt,align=center},
  gold/.style={nd,line width=1.1pt},
  pred/.style={nd,dashed},
  e/.style  ={-latex},
]
\node[nd]  (nl)  at (0,0)      {nature lieu};
\node[nd]  (oa)  at (10,0)     {objet architecture};
\node[nd]  (an)  at (0,-1.2)   {animaux};
\node[nd]  (aq)  at (-2.7,-2.4){animaux\\aquatiques};
\node[nd]  (te)  at (0.2,-2.4) {animaux\\terrestres};
\node[nd]  (oi)  at (2.9,-2.4) {oiseaux et\\volatiles};
\node[pred](ang) at (-3.7,-3.8){anguille};
\node[gold](sau) at (-1.9,-3.8){saumon};
\node[gold](her) at (0.2,-3.8) {h\'erisson};
\node[pred](aig) at (2.9,-3.8) {aigle};
\node[nd]  (sc)  at (10,-1.2)  {(sous-cat.)};
\node[gold](cha) at (10,-2.4)  {ch\^ateau};
\draw[e] (nl)--(an); \draw[e] (an)--(aq); \draw[e] (an)--(te); \draw[e] (an)--(oi);
\draw[e] (aq)--(ang); \draw[e] (aq)--(sau); \draw[e] (te)--(her); \draw[e] (oi)--(aig);
\draw[e] (oa)--(sc); \draw[e] (sc)--(cha);
\foreach \y/\d in {0/0,-1.2/1,-2.4/2,-3.8/3} \node[font=\tiny,gray] at (13,\y) {depth \d};
\end{tikzpicture}%
}
\caption{TIMEL excerpt for the worked example. Thick border = gold; dashed = predicted but not gold;
\emph{h\'erisson} and \emph{ch\^ateau} are both gold and predicted. The five TIMEL roots are
disjoint, so \emph{ch\^ateau} shares no ancestor with the animals.}
\label{fig:app-tree}
\end{figure}
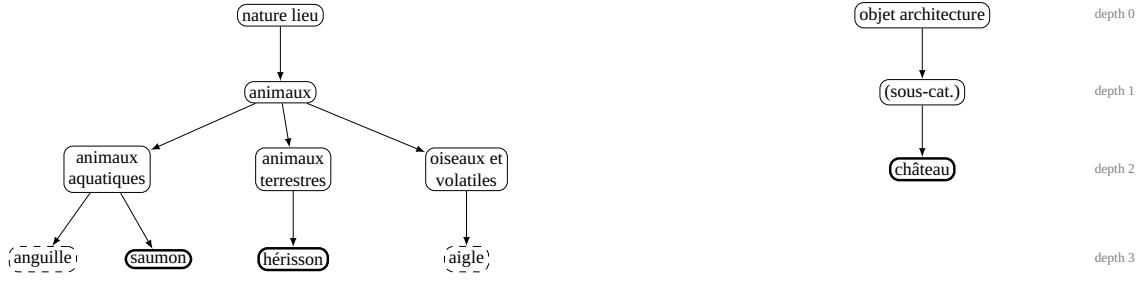

Consider an illumination whose gold annotation is
$G=\{\textnormal{saumon},\ \textnormal{h\'erisson},\ \textnormal{ch\^ateau}\}$ (a fish, a land
animal, a building). A model returns the ranked list
\[
P=[\ \textnormal{anguille},\ \textnormal{h\'erisson},\ \textnormal{aigle},\ \textnormal{ch\^ateau},\ \textnormal{``monstre marin''}\ ],
\]
in emission order, the last item being outside the TIMEL vocabulary. Figure~\ref{fig:app-tree} shows
the relevant taxonomy excerpt with the depths used below.

\paragraph{Output validity.} Of the five emitted labels, one (\emph{monstre marin}) matches no
canonical TIMEL leaf, so $\mathrm{invalid}=1/5=0.20$. The four valid predictions form the set
$P_v=\{\textnormal{anguille},\textnormal{h\'erisson},\textnormal{aigle},\textnormal{ch\^ateau}\}$.

\paragraph{Ranked (assistive) metrics.} Scanning the emitted list top-down (a correct leaf is one in
$G$):
\begin{center}\small
\begin{tabular}{clccccc}
\toprule
$k$ & item at rank $k$ & valid & in $G$ & $\mathrm{recall}@k$ & $\mathrm{precision}@k$ & $F_1@k$ \\
\midrule
1 & anguille       & yes & no  & 0.00 & 0.00 & 0.00 \\
2 & h\'erisson     & yes & yes & 0.33 & 0.50 & 0.40 \\
3 & aigle          & yes & no  & 0.33 & 0.33 & 0.33 \\
4 & ch\^ateau      & yes & yes & 0.67 & 0.50 & 0.57 \\
5 & monstre marin  & no  & no  & 0.67 & 0.40 & 0.50 \\
\bottomrule
\end{tabular}
\end{center}
Hence $\mathrm{precision}@1=0$ (the top suggestion, \emph{anguille}, is not gold), the first correct
leaf is at rank $2$ so $\mathrm{MRR}=1/2=0.5$, and at the list length $F_1@5=0.50$
($\mathrm{recall}@5=0.67$, $\mathrm{precision}@5=0.40$). On real data we use $K\in\{5,10,20\}$; here
the list is short, so we show $k=1\ldots5$.

\paragraph{Flat set metrics.} With $P_v$ and $G$ as sets,
$\mathrm{TP}=|P_v\cap G|=2$ (\emph{h\'erisson}, \emph{ch\^ateau}),
$\mathrm{FP}=|P_v\setminus G|=2$ (\emph{anguille}, \emph{aigle}),
$\mathrm{FN}=|G\setminus P_v|=1$ (\emph{saumon}); so $P=0.50$, $R=0.67$, $F_1=0.57$, and the
exact-match rate is $0$.

\paragraph{Soft metrics.} Using
$\mathrm{sim}_{\mathrm{soft}}(a,b)=\mathrm{depth}(\mathrm{LCA})/\max(\mathrm{depth}(a),\mathrm{depth}(b))$
(animals at depth $3$, \emph{ch\^ateau} at depth $2$):
\begin{center}\small
\begin{tabular}{lcccc}
\toprule
$\mathrm{sim}_{\mathrm{soft}}$ & saumon & h\'erisson & ch\^ateau & $\max_g$ \\
\midrule
anguille   & 0.67 & 0.33 & 0.00 & 0.67 \\
h\'erisson & 0.33 & 1.00 & 0.00 & 1.00 \\
aigle      & 0.33 & 0.33 & 0.00 & 0.33 \\
ch\^ateau  & 0.00 & 0.00 & 1.00 & 1.00 \\
\midrule
$\max_p$   & 0.67 & 1.00 & 1.00 & \\
\bottomrule
\end{tabular}
\end{center}
$P_{\mathrm{soft}}=\tfrac{1}{4}(0.67{+}1.00{+}0.33{+}1.00)=0.75$,
$R_{\mathrm{soft}}=\tfrac{1}{3}(0.67{+}1.00{+}1.00)=0.89$, so $F_{1,\mathrm{soft}}=0.81$. The soft
score rewards \emph{anguille} (a sibling of the gold \emph{saumon}, $0.67$) and \emph{aigle} (same
\emph{animaux} super-branch, $0.33$), credit that the flat $F_1$ ignores.

\paragraph{Hierarchical metrics.} Ancestor sets (root included):
$A(P_v)$ has $11$ nodes (the two animal roots-to-leaves plus \emph{objet architecture}$\to$sous-cat.
$\to$\emph{ch\^ateau}), $A(G)$ has $9$. Their intersection is
$\{$nature lieu, animaux, aquatiques, terrestres, h\'erisson, objet architecture, sous-cat.,
ch\^ateau$\}$, i.e.\ $\mathrm{TP}=8$; $\mathrm{FP}=3$ (anguille, oiseaux, aigle) and $\mathrm{FN}=1$
(saumon). Thus $H\text{-}P=8/11=0.73$, $H\text{-}R=8/9=0.89$, $H\text{-}F_1=0.80$. Note that
\emph{aigle} barely penalises the score: its ancestors \emph{nature lieu} and \emph{animaux} are
already shared with the gold, so only \emph{oiseaux} and \emph{aigle} count as false positives.

\paragraph{Reading.} On this instance the strict views are harsh (exact-match $0$,
$\mathrm{precision}@1=0$), the ranked view is moderate ($\mathrm{MRR}=0.5$, $F_1@5=0.50$, flat
$F_1=0.57$), and the hierarchy-aware views are high ($H\text{-}F_1=0.80$, $F_{1,\mathrm{soft}}=0.81$)
because the wrong leaves still sit in the correct branches; the IC-weighted soft $F_1$ would push
these up or down depending on how rare \emph{anguille}, \emph{h\'erisson} and \emph{ch\^ateau} are.
The invalid rate ($0.20$) separately flags the one hallucinated term.


\subsection{Training loss functions}
\label{app:losses}

\paragraph{Binary cross-entropy (BCE).}
For $C$~classes, binary gold vector $\mathbf{y}\in\{0,1\}^C$, and sigmoid predictions
$\hat{\mathbf{p}}\in(0,1)^C$:
\[
\mathcal{L}_{\mathrm{BCE}}
  = -\frac{1}{C}\sum_{c=1}^{C}
    \Bigl[y_c\log\hat{p}_c + (1-y_c)\log(1-\hat{p}_c)\Bigr].
\]
With $C{=}2{,}384$ and a mean of 8--12 positive labels per image, over 99\,\% of the gradient
signal originates from easy negatives ($y_c{=}0$, $\hat{p}_c{\approx}0$), which undermines
discrimination at the positive end of the score distribution.

\paragraph{Asymmetric Loss (ASL)~\cite{ridnik2021}.}
ASL decouples positive and negative contributions via independent focusing exponents and a
probability margin:
\[
\mathcal{L}_{\mathrm{ASL}}
  = -\sum_{c=1}^{C}
    \Bigl[
      y_c(1-\hat{p}_c)^{\gamma^+}\log\hat{p}_c
      +(1-y_c)\,\hat{p}_{c,m}^{\gamma^-}\log(1-\hat{p}_{c,m})
    \Bigr],
\]
where $\hat{p}_{c,m}=\max(\hat{p}_c - m,\,0)$.
Setting $\gamma^+{=}0$ leaves positive gradients unattenuated; $\gamma^-{=}4$ strongly
downweights easy negatives; $m{=}0.05$ hard-zeros negligible negative probabilities before
the focal term is applied. These values are used in all ASL runs.

\paragraph{Hierarchical cross-entropy (hier-CE).}
The CAT decoder accumulates a cross-entropy at each active level $\ell$ of the taxonomy
traversal:
\[
\mathcal{L}_{\mathrm{hier\text{-}CE}}
  = -\sum_{\ell}\sum_{c\in\mathcal{C}(\ell)}
    y_c^{(\ell)}\log p(c_\ell\mid c_{<\ell},x),
\]
where $\mathcal{C}(\ell)$ is the set of valid children at level~$\ell$, $y_c^{(\ell)}$ is the
binary gold indicator for concept~$c$, and $p(c_\ell\mid c_{<\ell},x)$ is the conditional
softmax probability given the partial path decoded so far. Supervision at intermediate nodes
provides gradient signal to all ancestors of a leaf, benefiting rare leaves sharing parents with
more frequent siblings.

\paragraph{Hierarchical ASL (hier-ASL).}
The hier-ASL variant substitutes ASL for the per-level cross-entropy within the CAT framework,
retaining the class-imbalance correction of ASL inside the tree traversal. Formally, for each
level $\ell$ with children $\mathcal{C}(\ell)$:
\[
\mathcal{L}_{\mathrm{hier\text{-}ASL}}
  = \sum_\ell \mathcal{L}_{\mathrm{ASL}}\!\left(
      \sigma\!\left(W^{(\ell)}\mathbf{z}^{(\ell)}\right),\;
      \mathbf{y}^{(\ell)}
    \right),
\]
with $\gamma^+{=}0$, $\gamma^-{=}4$, $m{=}0.05$.


\printbibliography

\section*{Usage of AI}

The authors used OpenAI's ChatGPT as an AI-assisted writing support tool for the iterative refinement of the manuscript, including improvements in wording, organization, and translation. The scientific content, analyses, and conclusions remain the sole responsibility of the authors.

\end{document}